\documentclass[12pt]{article}
\usepackage[margin=1in]{geometry}

\usepackage{authblk}

\usepackage{natbib}
\setcitestyle{authoryear}

\usepackage{hyperref}

\usepackage{xcolor}

\newcommand{\diff}{{\rm d}}
\newcommand{\calA}{\mathcal{A}}
\newcommand{\calB}{\mathcal{B}}
\newcommand{\calN}{\mathcal{N}}
\newcommand{\calX}{\mathcal{X}}
\newcommand{\R}{\mathbb{R}}

\usepackage{amsfonts,amsmath,graphicx,makecell}
\usepackage{multirow}

\begin{document}

\title{Improving Lunar Topography with Deep Learning Schr{\"o}dinger Bridges}

\author[1]{Matthew Repasky\thanks{Email: mwrepasky@gmail.com}}
\author[2]{Erwan Mazarico} 
\author[2]{Michael K. Barker}
\author[2,3,4]{Stefano Bertone}
\author[2]{Terence J. Sabaka}
\author[1]{Yao Xie}

\affil[1]{
\small
H. Milton Stewart School of Industrial and Systems Engineering, Georgia Institute of Technology, USA}
\affil[2]{
\small
NASA Goddard Space Flight Center, USA
}
\affil[3]{
\small
Center for Research and Exploration in Space Science and Technology (CRESST II), University of Maryland, College Park, MD, USA
}
\affil[4]{
\small
National Institute for Astrophysics (INAF), Astrophysical Observatory of Turin, 10025 Pino Torinese (TO), Italy
}

\date{}

\maketitle

\noindent\textcolor{red}{This is the version of the article before peer review or editing, as submitted by an author to The Planetary Science Journal. IOP Publishing Ltd is not responsible for any errors or omissions in this version of the manuscript or any version derived from it. The Version of Record is available online at \href{https://doi.org/10.3847/PSJ/ae6244}{10.3847/PSJ/ae6244} (Matthew Repasky \textit{et al} 2026 \textit{Planet. Sci. J.} \textbf{7} 139)
}
 
\begin{abstract}

Increasing the resolution of planetary topography models can enable a better understanding of surface processes {and geomorphology}; however, existing analytical super-resolution methods are expensive and difficult to apply at large scales. Generative models provide the tools to learn complex relationships within data and can be applied at scale due to hardware accelerators and parallelization. We present a diffusion-based Schr{\"o}dinger Bridge (SB) generative modeling approach for lunar topography super-resolution, connecting the distribution of low-resolution topography to that of high-resolution topography, incorporating physically-constraining optical imagery. Our approach is inspired by existing Shape-from-Shading methods, which improve \textit{a priori} low-resolution topography by using optical images at the target resolution. We train SBs on a novel dataset of rendered lunar topography, emulating optical imagery from the Lunar Reconnaissance Orbiter Narrow Angle Camera. The result is a flexible approach for topography super-resolution which can provide pixel-level uncertainties in the reconstruction.

\end{abstract}

\textit{Unified Astronomy Thesaurus concepts:} {{Lunar surface} --- {Lunar science} --- {The Moon} --- {Neural networks} --- {Remote sensing}}

\section{Introduction}

Topography models of planetary surfaces derived from orbiter measurements provide information about surface processes and the geological history of planetary bodies. For example, Digital Elevation Models (DEMs) based upon measurements made by the Lunar Orbiter Laser Altimeter (LOLA) onboard the Lunar Reconnaissance Orbiter (LRO) represent topography at varying resolution across the lunar surface~\citep{barker2016new,barker2023new}. However, their limited resolution inhibits the analysis of small-scale surface features, which could provide insight into current surface processes such as mass wasting and micrometeorite bombardment. Moreover, characterization of landing-site topography at scales relevant to human and robotic exploration can support NASA's Artemis lunar exploration missions~\citep{barker2021improved}. Analytical methods for improving the resolution of \textit{a priori} DEMs typically incorporate additional data sources, such as optical imagery. For instance, the LRO Narrow Angle Camera (NAC) and Korea Pathfinder Lunar Orbiter (KPLO) ShadowCam resolve the lunar surface down to 1-3 meters-per-pixel (mpp)~\citep{robinson2010lunar,robinson2017shadowcam}. Shape-from-shading (SfS) refers to a class of methods utilizing brightness information from optical imagery~\citep[see, e.g.,][]{alexandrov2018multiview,fernandes2022high}, to yield DEMs of higher resolution than those created using stereophotogrammetry~\citep{henriksen2017extracting}. However, they are relatively inflexible, requiring expert-tuned parameters and precise specification of illumination and viewing angles. Additionally, these methods are expensive, hindering their application to large areas of interest for analysis and exploration.

In computer vision, generative models have been applied to solve inverse problems such as Super-Resolution (SR), {which is concerned with increasing the pixel resolution of array inputs.} Diffusion models, in particular, have recently arisen as a powerful class of generative model~\citep{ho2020denoising,song2020score}. Such models can be leveraged for conditional image synthesis techniques such as guidance of the generative process using class labels~\citep{song2020score,dhariwal2021diffusion} or infusion of the conditional data into the neural network~\citep{rombach2022high}. {Generally, conditional models take contextual information into account when synthesizing data.} Image SR diffusion models~\citep{saharia2022image,saharia2022palette} take the degraded data as input into the diffusion model neural network to yield high-resolution output, outperforming conditional synthesis approaches based upon predecessor generative models such as Generative Adversarial Networks (GANs)~\citep{goodfellow2014generative}. Despite their advantages, typical diffusion models initialize the generative process with random noise, which may be inefficient. The image-to-image Schr{\"o}dinger bridge (SB) method is a generative modeling approach whose training and generation procedures are based on diffusion; however, they initialize synthesis with the \textit{a priori} (degraded) data~\citep{liu2023}. In image reconstruction tasks such as SR, SBs yield demonstrably better reconstructions than those synthesized via diffusion~\citep{liu2023}.

\begin{figure}[t!]
    \centering
    \includegraphics[width=0.8\textwidth]{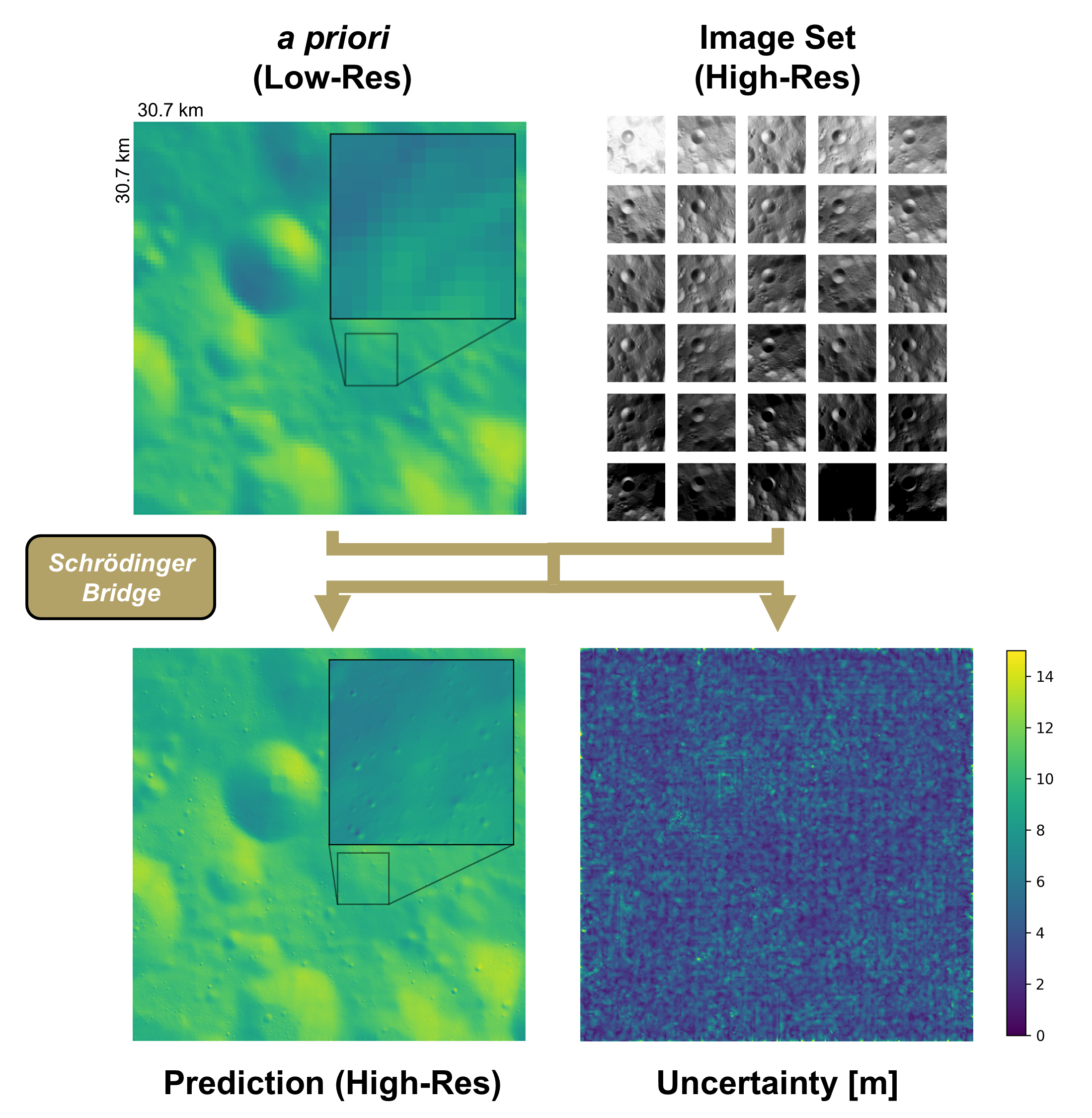}
    \caption{The SB method for topography SR transforms \textit{a priori} low-resolution topography into high-resolution reconstructions, taking a set of high-resolution images as additional conditional input. Multiple samples from the SB model can be used to create uncertainty maps.}
    \label{fig:main_method}
\end{figure}

Due to the high dimension and complexity of remote sensing data, deep learning and computer vision methods have been applied to many geophysical problems. {Generally, these methods avoid user-tunable parameters via data-driven function approximation, trading off up-front computation (training neural networks) for rapid inference (producing high-resolution data products).} For example, prior works have applied deep generative models to SR of satellite optical image data using GANs~\citep{hu2024super,delgado2023superresolution}, normalizing flows~\citep{heintz2023multiscale}, and diffusion~\citep{illarionova2023benchmark}. \cite{delgado2023superresolution} developed a GAN approach for the SR of NAC imagery, which can then be used for downstream tasks such as building topography models. Other works have directly applied similar methods for SR of planetary topography models. In particular, GANs have been applied for topography SR on Mars, constraining high-resolution topography using optical imagery~\citep{tao2021single,tao2021madnet,tao2022subpixel}. However, the generative modeling literature suggests that diffusion-based approaches outperform GANs in both unconditional synthesis~\citep{dhariwal2021diffusion} and in conditional generation tasks, such as those related to inverse problems~\citep{chung2022diffusion,liu2023}. Cutting-edge diffusion-based generative modeling techniques such as SBs have yet to be applied directly to topography SR.

In this work, {we develop a machine learning method to increase the resolution of existing topography models when supplying the algorithm with a low-resolution \textit{a priori} DEM and multiple corresponding optical images; see Figure~\ref{fig:main_method} for a visualization.} We propose a topography SR method based on the SB framework{, advancing previous AI-based approaches for topography improvement}. Inspired by SfS (see \cite{alexandrov2018multiview} and references therein), deep learning methods for topography super-resolution \citep{tao2022subpixel}, and the superior performance of SBs~\citep{liu2023}, we define a conditional SB model which transforms a low-resolution, \textit{a priori} DEM into a high-resolution counterpart given a set of target-resolution optical images as context. Intuitively, initializing the generative process with the low-resolution measurement is potentially more efficient than initializing it with pure noise, as is done in typical diffusion models~\citep{ho2020denoising} and normalizing flows~\citep{lipman2022flow}. To develop and demonstrate our methodological approach, we work in a simulated setting with a preliminary evaluation of real data. We train SBs to recover LOLA DEMs at 20 mpp given \textit{a priori} topography at 320 mpp and sets of rendered optical images emulating NAC images at 20 mpp. {We emphasize that the goal of this work is to define and implement a new approach to topography SR based on some of the latest deep-learning methods. We show that, at this stage of experimentation, we obtain promising results, but that further work will be required to make our method applicable to more general use in planetary science and more directly compete with established non-machine-learning SfS approaches.}

{This manuscript is organized as follows. In Section~\ref{sec:method}, we give a detailed background and assessment of deep learning methods relevant to developing our chosen methodology. In Section~\ref{sec:datasets}, we show we build the necessary training dataset, from existing lunar topography models and for realistic illumination and observing conditions. Section~\ref{sec:experiment} gives details on the implementation of our model, its training, and evaluation. In Section~\ref{sec:application}, we apply our trained model to large-scale topography reconstruction from imagery under increasingly realistic conditions. Finally, in Section~\ref{sec:conclusions}, we summarize our findings.}

\begin{figure}[t!]
    \centering
    \includegraphics[width=\linewidth]{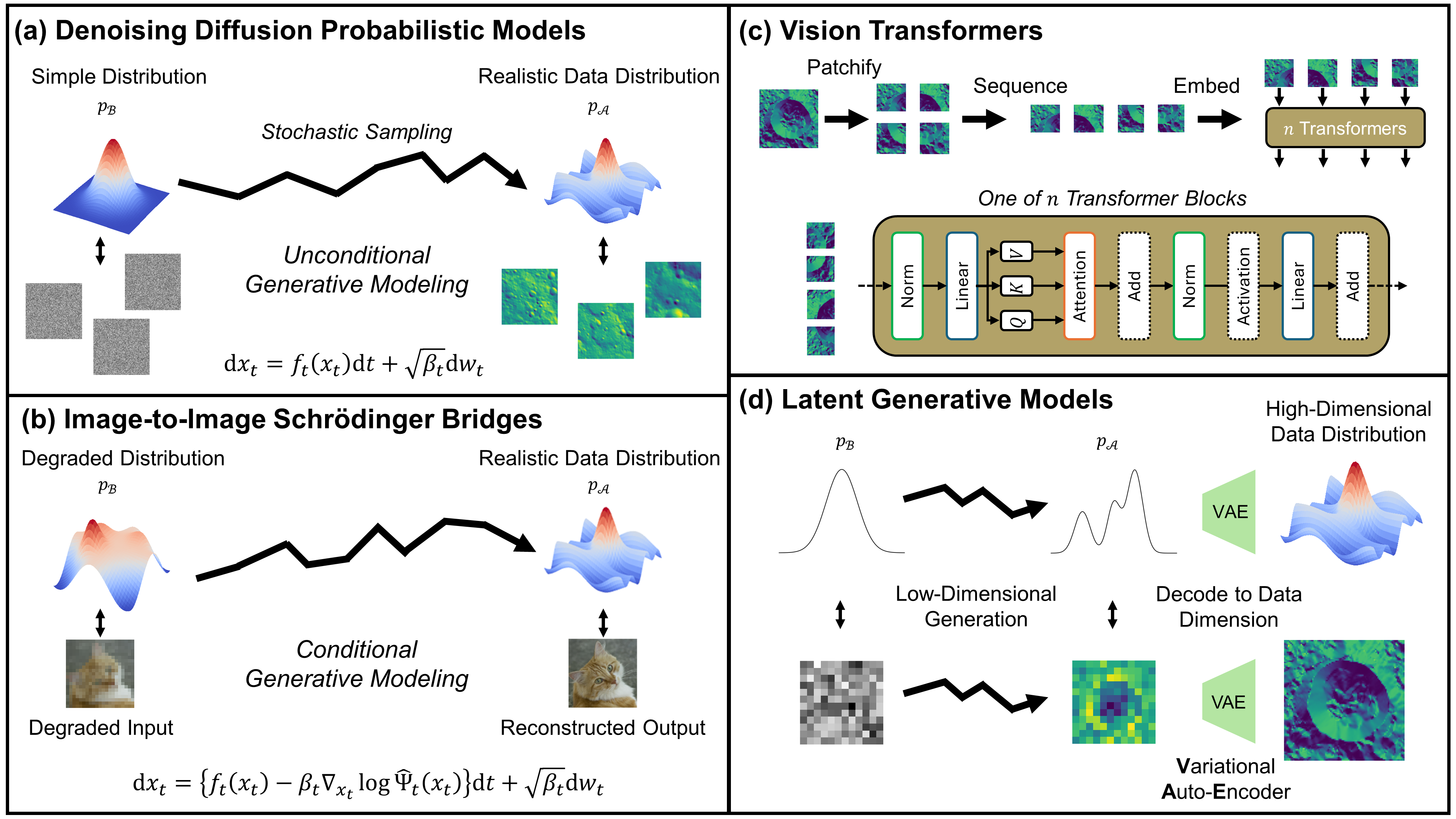}
    \caption{Diffusion models (a) link a data-generating distribution {(high-resolution DEMs)} to the Gaussian distribution {(white noise)} via an SDE, while SBs (b) link two arbitrarily complicated distributions. (c) The reverse-time dynamics of these samplers can be modeled using vision transformers. Representing these models in the latent space of a VAE (d) is often more efficient.}
    \label{fig:background}
\end{figure}

{
\textbf{Notation.} For brevity, we denote time dependence of functions $f(\cdot,t)$ using a subscript $f_t$. Let $\calX$ denote a data manifold, for instance, $\calX=\R^d$ or $\calX=\R^{d\times d}$. We define $\Omega=C([0,T],\calX)$ as the set of all continuous $\calX$-valued functions on the \textit{time interval} $[0,T]$. A \textit{path measure} is defined as any positive measure $\mathbb{P}$ on $\Omega$~\citep{leonard2013survey}. Given time-marginal measures $\mathbb{P}_t$ as the push-forward of $\mathbb{P}$ to a time $t\in[0,T]$, the associated marginal probability density is denoted $p_t(x_t)$ for $x_t\in\calX$, i.e., $\diff\mathbb{P}_t = p_t \diff x_t$. Superscripts $\rm{DM}$ and $\rm{SB}$ are used to distinguish between measures and densities associated with Diffusion Models and Schr{\"o}dinger Bridges, respectively; see Section~\ref{sec:method} for details. Finally, we denote the Laplacian by $\Delta_{x} f=\nabla_{x} \cdot \nabla_x f$.
}

\section{Topography Super-Resolution \& Generative Models}\label{sec:method}

{In this work, we formulate the task of increasing the resolution of planetary topography models in the framework of imaging super-resolution (SR), which in machine learning refers to a category of methods for increasing the resolution of raster array dataset~\citep{nasrollahi2014super,anwar2020deep}. Moreover, following modern deep learning SR approaches, we formulate this task as posterior sampling~\citep{chung2022diffusion}}.

Specifically, let $x_T$ represent an \textit{a priori} low-resolution DEM. SR can be posed as conditional sampling:
\begin{equation}\label{eq:sr_problem}
    x_0 \sim p_\calA(\cdot | x_T, \mathcal{Y}),
\end{equation}
where $x_0$ is a high-resolution patch corresponding to {the same spatial region as} $x_T$ and $\mathcal{Y}$ contains additional contextual information about $x_0$ (e.g., $\mathcal{Y}:=\{ y_j\}_{j=1}^J$). {Equation~\ref{eq:sr_problem} suggests that the high-resolution DEM is randomly distributed according to $p_{\calA}$, a distribution conditioned on $x_T$ and $\mathcal{Y}$.} This framework is inspired by SfS, which increases the resolution of topography provided high-resolution image context.

{
Topography and imagery are typically represented by high-dimensional raster arrays {(i.e., arrays with a large number of pixels)}. Therefore, approximating the posterior (Equation~\ref{eq:sr_problem}) via classical and analytical methods is intractable for moderately-sized problems. This is further complicated by the fact that the parametric form of $p_{\calA}$ is unknown and complicated in general. Hence, modern computer vision techniques for SR rely on deep learning techniques. The probabilistic formulation of Equation~\ref{eq:sr_problem} naturally lends itself to tools developed within the evolving field of generative modeling, and such techniques are commonly applied to modern SR problems~\citep{saharia2022image,saharia2022palette,chung2022diffusion}. Modern deep generative models \citep{song2020score,lipman2022flow,liu2023} act as samplers from intractable data distributions, and can be conditioned on added contextual information. In this work, we apply generative modeling to approximate the sampling of Equation~\ref{eq:sr_problem}.
}

The following sub-sections provide a brief review of relevant generative modeling frameworks, followed by an overview of neural network function approximators typically applied for learning generative models. Modern methods define a path measure $\mathbb{P}$ interpolating between $p_\calA$ and another probability density $p_\calB$. In Section~\ref{sec:diffusion}, we described how diffusion models define $p_\calB$ as an analytically tractable density, e.g., a centered Gaussian $\calN(0,I)$. SBs, defined in Section~\ref{sec:sb_prob}, generalize diffusion such that $p_\calB$ is also a complicated, data-generating density (an \textit{a priori} distribution). The generative process is a data dynamic $x_t$ for $t\in[0,T]$ transforming $x_T\sim p_\calB$ into a sample $\hat{x}_0$ distributed approximately according to $p_\calA$, the ``target" distribution. {This allows one to more directly persist information from the \textit{a priori} data to recover high-resolution topography.} {Section~\ref{sec:sb} details how SB can be applied to diffusion-like problems, and Section~\ref{sec:topo_sr_sb} presents how, in this work, we propose a SB problem to approximate posterior sampling via Equation~\ref{eq:sr_problem}.}

\subsection{Denoising Diffusion Probabilistic Models}\label{sec:diffusion}

Diffusion models~\citep{ho2020denoising,song2020score} are defined by a system of forward- and reverse-time stochastic differential equations (SDEs) that share marginal densities $p_t^{\rm DM}$~\citep{anderson1982reverse}:
\begin{equation}\label{eq:diffusion_sde_forward}
    \diff x_t = f_t(x_t) \diff t + \sqrt{\beta_t} \diff w_t, \quad x_0\sim p_\calA,
\end{equation}
\begin{equation}\label{eq:diffusion_sde_reverse}
    \diff x_t = \left\{ f_t(x_t) - \beta_t\nabla_{x_t} \log p_t^{\rm DM}(x_t) \right\} \diff t + \sqrt{\beta_t} \diff \bar{w}_t, \quad x_T\sim p_\calB,
\end{equation}
where $f_t$ is the time-varying drift, $\beta_t$ is a user-defined variance schedule, and $\diff w_t$/$\diff \bar{w}_t$ is an infinitesimal Wiener process in forward/reverse time. For appropriately specified $\beta_t$ and $f_t$ linear in $x_t$, Equation~\ref{eq:diffusion_sde_forward} simulates $p_\calB:=\calN(0,I)$ by large enough time $t=T$ and Equation~\ref{eq:diffusion_sde_reverse} simulates $p_\calA$ starting from Gaussian noise at time $T$. The unconditional generative process defined by diffusion models is visualized in Figure~\ref{fig:background}(a).

Deep generative models based on diffusion approximate the score function $\nabla_{x_t}\log p_t^{\rm DM}(x_t)$ using a neural network $u^\theta_t(x_t)$ parameterized by $\theta$. Given $u^\theta_t(x_t)\approx\nabla_{x_t}\log p_t^{\rm DM}(x_t)$, Equation~\ref{eq:diffusion_sde_reverse} can be numerically simulated to produce samples from $p_\calA$. In practice, this approximation is conducted using an objective function resembling Denoising Score Matching (DSM)~\citep{vincent2011connection}:
\begin{equation}\label{eq:dsm}
    \theta^\star = \underset{\theta}{\text{arg min}} \ \mathbb{E} \ \| u_t^\theta(x_t) - \sigma_t\nabla_{x_t} \log p_t^{\rm DM}(x_t | x_0) \|,
\end{equation}
where $\sigma^2_t=\int_0^t\beta_\tau \diff\tau$ {and the expected value is over $x_0\sim p_\calA$, $t\sim\mathcal{U}(0,T)$, and $x_t\sim p_t^{\rm DM}(\cdot|x_0)$}. This objective has minimizer $u_t^{\theta^\star}(x_t)\approx\nabla_{x_t}\log p_t^{\rm DM}(x_t)$. The conditional score can be computed analytically, as $\nabla_{x_t} \log p_t^{\rm DM}(x_t | x_0)=(x_t-x_0)/\sigma_t^2$, since the forward-process conditional density has a known Gaussian closed form~\citep{song2019generative,ho2020denoising}, which makes the objective of Equation~\ref{eq:dsm} tractable. Given $u_t^{\theta^\star}$ and some $x_T\sim p_\calB$, Equation~\ref{eq:diffusion_sde_reverse} can be simulated via recursive posterior sampling, otherwise known as ancestral sampling~\citep{song2020score}, $x_{t-1} \sim p^\theta(x_{t-1}|x_t)$:
\begin{equation}\label{eq:ancestral_sampling}
    x_{t-1} = \frac{1}{\sqrt{1-\beta_t}} \left( x_t + \beta_t u_t^{\theta^\star}(x_t) \right) + \sqrt{\beta_t}\xi, \quad \xi\sim\calN(0,I).
\end{equation}

\subsection{The Schr{\"o}dinger Bridge Problem}\label{sec:sb_prob}
{
The SB problem is concerned with identifying a path measure connecting two arbitrary distributions~\citep{schrodinger1932theorie,leonard2013survey}. {In this work, these distributions represent the low- and high-resolution topography, and the path measure is an interpolation between them.} This can be expressed as follows~\citep{pavon1991free,chen2021likelihood}:
}
\begin{equation}\label{eq:sb_problem}
{
    \mathbb{Q}^\star = \underset{\mathbb{Q}}{\text{arg min}} \ D_{\rm KL} (\mathbb{Q} \| \mathbb{P}) \quad\text{such that}\quad \diff\mathbb{P}_0 = p_\calA \diff x_0, \ \diff\mathbb{P}_T = p_\calB \diff x_T. 
}
\end{equation}
{
That is, given some ``reference measure'' $\mathbb{P}$, the objective is the closest path measure $\mathbb{Q}^\star$ to $\mathbb{P}$ which has $p_\calA$ and $p_\calB$ as initial and terminal marginal densities, respectively.
}

{
When the reference measure $\mathbb{P}$ is the path measure of the diffusion SDEs (Equations \ref{eq:diffusion_sde_forward} and \ref{eq:diffusion_sde_reverse}), the solution $\mathbb{Q}^\star$ to Equation~\ref{eq:sb_problem} can be determined using the \textit{potentials} $\Psi_t$ and $\hat{\Psi}_t$, which are the solutions to the following system of Partial Differential Equations (PDEs)~\citep{caluya2021wasserstein,chen2021likelihood}:
}
\begin{equation}\label{eq:sb_pde}
{
    \begin{aligned}
        \frac{\partial\Psi_t}{\partial t} &= -\left( \nabla_{x_t}\Psi_t \right) \cdot f_t - \frac{\beta_t}{2} \Delta_{x_t}\Psi_t, \\
        \frac{\partial\hat{\Psi}_t}{\partial t} &= -\nabla_{x_t} \cdot \left( \hat{\Psi}_t f_t \right) + \frac{\beta_t}{2} \Delta_{x_t}\hat{\Psi}_t, \\
        \text{such that }& \quad \Psi_0\hat{\Psi}_0=p_\calA, \quad \Psi_T\hat{\Psi}_T=p_\calB.
    \end{aligned}
}
\end{equation}
{
The marginal density of the optimal path measure $\mathbb{Q}^\star$ is $p_t^{\rm SB}(x_t)=\Psi_t(x_t)\hat{\Psi}_t(x_t)$ for every $t\in[0,T]$. Moreover, this path measure and the associated marginal densities correspond to the following system of SDEs:
}
\begin{equation}\label{eq:sb_sde_forward}
    \diff x_t = \left\{ f_t(x_t) + \beta_t \nabla_{x_t} \log \Psi_t(x_t) \right\} \diff t + \sqrt{\beta_t}\diff w_t,\quad x_0\sim p_\calA,
\end{equation}
\begin{equation}\label{eq:sb_sde_reverse}
    \diff x_t = \left\{ f_t(x_t) - \beta_t \nabla_{x_t} \log \hat{\Psi}_t(x_t) \right\} \diff t + \sqrt{\beta_t}\diff\bar{w}_t,\quad x_T\sim p_\calB.
\end{equation}

{
Similar to the approach taken with diffusion models, Equation~\ref{eq:sb_sde_reverse} can be simulated in reverse time to transform samples $x_T\sim p_\calB$ into corresponding samples from $p_\calA$. However, the key difference is that $p_\calB$ can be generalized to an arbitrarily complicated \textit{a priori} distribution. The SB generative process linking two complicated distributions is visualized in Figure~\ref{fig:background}(b). To simulate the generative process, $\Psi_t$ and $\hat{\Psi}_t$ need to be obtained as solutions to the system of PDEs. There exist many methods to approximate these potentials, including segmentation of the bridge into many portions that can be solved analytically~\citep{wang2021deep}, using iterative proportional fitting methods~\citep{de2021diffusion,vargas2021solving}, or via forward-backward SDEs theory~\citep{chen2021likelihood}.
}

\subsection{Image-to-Image Schr{\"o}dinger Bridges}\label{sec:sb}

{Image-to-Image SBs~\citep{liu2023} approximate the SB potential $\hat{\Psi}_t$ using an approach similar to the approach used to learn diffusion models~\citep{song2020score}. However, diffusion modeling techniques cannot immediately be applied to learn the dynamics of SBs due to the nonlinearity of the drift terms in the SDE Equation~\ref{eq:sb_sde_forward}.} \cite{liu2023} consider the tractable, linear SDE:
\begin{equation}\label{eq:sb_sde_forward_linear}
    \diff x_t = f_t(x_t) \diff t + \sqrt{\beta_t}\diff w_t,\quad x_0\sim\hat{\Psi}_0,
\end{equation}
whose limiting distribution is Gaussian. Analogous to the diffusion forward-time SDE (Equation~\ref{eq:diffusion_sde_forward}), Equation~\ref{eq:sb_sde_forward_linear} shares marginal densities with a reverse-time SDE with drift $f_t-\beta_t\nabla_{x_t}\log\hat{\Psi}_t$: the SDE of Equation~\ref{eq:sb_sde_reverse}. {Given $x_0\sim\hat{\Psi}_0$}, conditional densities $\hat{\Psi}(x_t|x_0)$ of Equation~\ref{eq:sb_sde_forward_linear} are analytically tractable. This suggests that diffusion modeling techniques can be employed to learn $u_t^{\theta^\star}(x_t)\approx\nabla_{x_t}\log\hat{\Psi}_t(x_t)$ {via a DSM-like objective function}:
\begin{equation}\label{eq:sb_dsm}
{
    \theta^\star = \underset{\theta}{\text{arg min}} \ \mathbb{E} \ \| u_t^\theta(x_t) - \sigma_t\nabla_{x_t} \log p_t^{\rm DM}(x_t | x_0) \|,
}
\end{equation}
{
where the expected value is over $x_0\sim p_\calA$, $x_T\sim p_\calB(\cdot|x_0)$, $t\sim\mathcal{U}(0,T)$, and $x_t\sim p_t^{\rm SB}(\cdot|x_0,x_T)$.
}

{However, two computational challenges must be addressed: (i) sampling $x_0\sim\hat{\Psi}_0$ is not tractable in general and (ii) the path measure induced by Equation~\ref{eq:sb_sde_forward_linear} is not the same as that induced by Equations~\ref{eq:sb_sde_forward} and \ref{eq:sb_sde_reverse}, i.e., $\mathbb{Q}^\star$ of Equation~\ref{eq:sb_problem}. To address issue (i), \cite{liu2023} define $p_\calA:=\delta_a(x)$, the Dirac delta distribution centered at $a\in\calX$. This eliminates one of the couplings in the constraints of Equation~\ref{eq:sb_pde}, resulting in $\hat{\Psi}_0=\delta_a$, from which samples ($a\sim\delta_a$) can be produced trivially.}
{To consider (ii), note that one cannot simply sample $x_t\sim p_t^{\rm DM}(x_t|x_0)$ induced by Equation~\ref{eq:sb_sde_forward_linear} when computing the DSM objective (Equation~\ref{eq:sb_dsm}), as this will not accurately represent the path between $p_\calA$ and $p_\calB$ (instead, it corresponds to a path between $p_\calA$ and a Gaussian).} To address this, \cite{liu2023} derive an analytic posterior $p^{\rm SB}(x_t|x_0,x_T)$ for boundary pair $x_0\sim p_\calA$ and $x_T\sim p_\calB$, which they prove is equivalent to the marginal form of the diffusion recursive posterior density $p(x_t|x_{t+1})$ of Equation~\ref{eq:ancestral_sampling}. Therefore, when training, $x_t$ can be sampled directly from $p^{\rm SB}(x_t|x_0,x_T)$ given boundary pairs. For sampling, $x_T\sim p_\calB(x_T|x_0)$ can be input to ancestral sampling to yield corresponding samples $x_0\sim p_\calA(x_0)$.

{
To summarize, the generative dynamics of the Image-to-Image SB can be modeled by a neural network via the minimization of Equation~\ref{eq:sb_dsm}.
}

\subsection{Topography Super-Resolution Schr{\"o}dinger Bridges}\label{sec:topo_sr_sb}
We train a SB model for 16$\times$ {super-resolution} (320 mpp to 20 mpp) of lunar DEMs. Let $x_0\in\mathbb{R}^{96\times96}$ and $x_T\in\mathbb{R}^{96\times96}$ represent high- and low-resolution DEMs, respectively, and let the contextual information be a (variable-size $J$) set of 20 mpp optical images: $\mathcal{Y}:=\{ y_j\}_{j=1}^J$ where $y_j\in\mathbb{R}^{96\times96}$. We define an SB connecting $x_T\sim p_\calB$ to $x_0\sim p_\calA$, conditioned on $\mathcal{Y}$. {Due to the high-dimension {(number of pixels)} and complex structure of $x_0$, $x_T$, and $\mathcal{Y}$, we optimize the parameters $\theta$ of a deep neural network $u^{\theta}_t$ representing the SB dynamics using Equation~\ref{eq:sb_dsm}, which is common practice for generative modeling and SR~\citep{song2020score,saharia2022palette}.}

{
Transformers represent the state-of-the-art in deep neural networks, which utilize the ``attention'' mechanism to relate disparate features in high-dimensional, structured data such as text~\citep{vaswani2017attention}. Such architectures have been adapted to vision generative modeling tasks in recent years~\citep{dosovitskiy2020image,peebles2023scalable}, dubbed vision transformers (ViTs) and diffusion transformers (DiTs). Additionally, many modern generative models mitigate the computational expense of modeling in the high-dimensional data space by defining a utility \textit{latent space}, represented by a variational autoencoder (VAE)~\citep{kingma2013auto,rombach2022high}.
}
In this work, we train a SB model in the latent space of a VAE, and the generative dynamics of the SB are modeled by a DiT.
{
See Appendix~\ref{sec:depp_learning} for an in-depth background of neural network models and autoencoders used in this work, and see Section~\ref{sec:parameterization} for the specific implementation for lunar topography SR.
}


{
{Given a trained lunar topography SB, high-resolution topography products can be created using few function evaluations of the trained network. Moreover, application of this model requires minimal user fine-tuning or specification, as surface conditions such as illumination and viewing angles are not required as input, unlike many traditional approaches.} Training deep neural networks to model SB dynamics via Equation~\ref{eq:sb_dsm} requires a large training dataset. In the following section, we outline this training dataset, which utilizes high-resolution topography data from the lunar surface, and rendered optical images emulating NAC imagery.
}

\begin{figure}[!t]
    \centering
    \includegraphics[width=1.\linewidth]{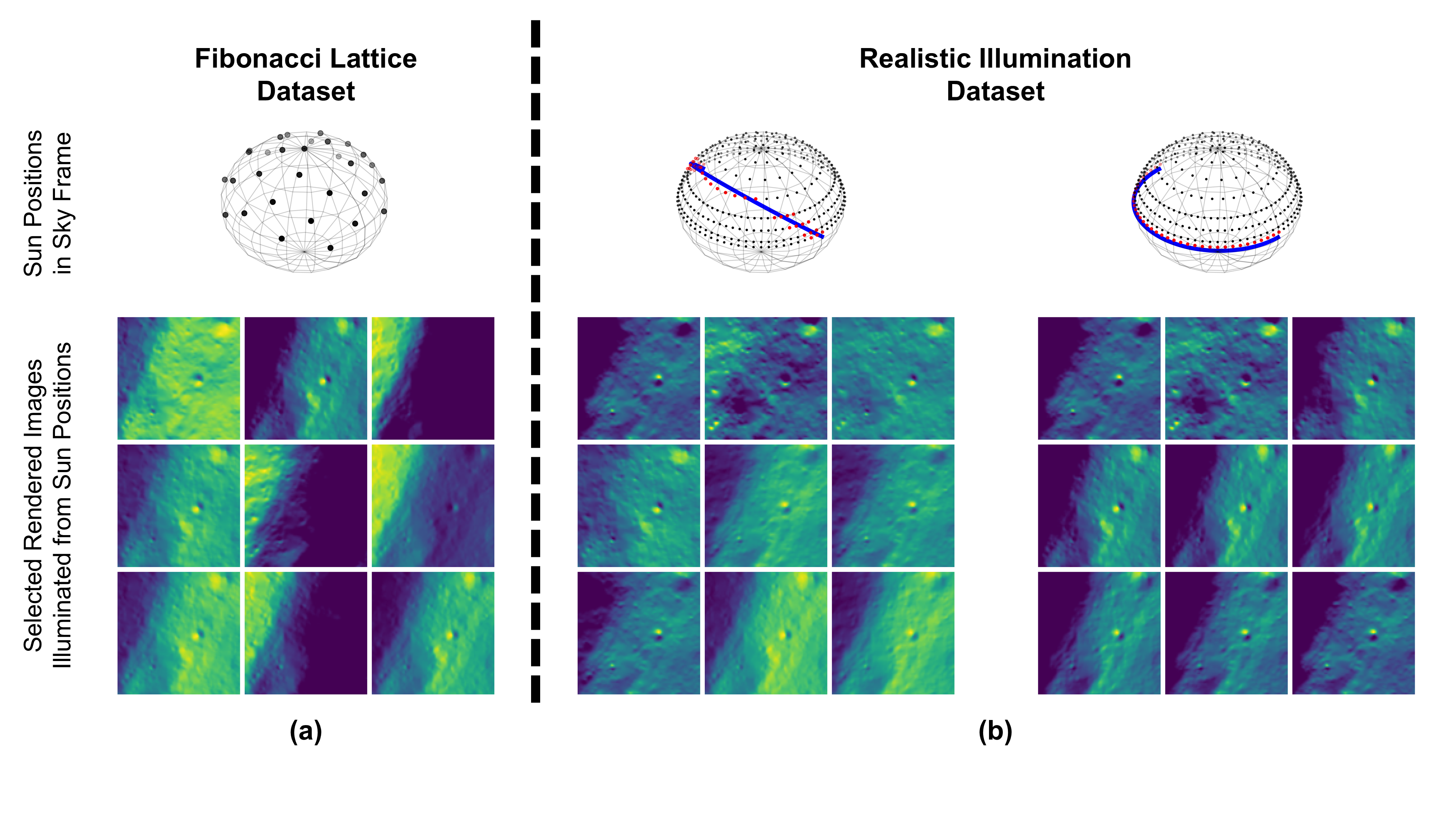}
    \caption{In (a), illumination directions are defined using a Fibonacci lattice across the sky to represent the entire spectrum of illumination conditions. More realistic illumination (b) is considered by randomly determining a location of the DEM patch on the lunar surface, corresponding to an arc of possible illumination directions. Rendered images with illumination directions near this arc are chosen as ``realistic''. In both columns of (b), the assumed longitude of the patch is 0\textdegree \ E. The left column assumes the latitude is 40\textdegree \ N while the right assumes 88\textdegree \ N.}
    \label{fig:dataset_illumination}
\end{figure}

\section{Rendered Lunar Topography Dataset}\label{sec:datasets}
This work applies deep generative SR techniques to lunar topography data. Due to the high quality of available topography data products near the poles at $\ge$ 20 m resolution, we focus on 20 mpp LOLA DEMs corresponding to both polar regions on the Moon polewards of 80{\textdegree} latitude~\citep{barker2023new,barker2025large}. To create a dataset for training deep SR models, the DEMs are cut into $96\times96$-pixel patches ($1.92\times1.92$ km) representing the target (``true") high-resolution topography. Due to variable LOLA altimetric data density coverage across the lunar surface, patches with few ($<$2,242, or $\sim$24\%) altimetry-defined (non-interpolated) pixels are discarded. The result is a dataset of 91,708 patches, which is split into an 80\%/10\%/10\% partition for model training, validation, and testing, respectively. 

We generate multiple rendered images of every DEM patch, to associate each with a set of simulated optical measurements {(i.e., the contextual information)}. This is conducted using raytracing-based illumination software~\citep{shadowspy} for rendering shape models of planetary topography~\citep{mazarico2011illumination,potter2023fast,mazarico2018advanced}. We create two rendered datasets corresponding to the set of 91,708 DEM patches, which are described in the subsequent sub-sections. Each dataset defines the positions of the Sun in the sky for the set of images in a different manner. Individual rendered images from these datasets are meant to emulate NAC-like imagery to be provided as contextual information for SR models of lunar topography.

\subsection{Fibonacci Lattice Illumination Dataset}\label{sec:fibonacci}
For each DEM patch, 30 Sun positions are specified via a Fibonacci lattice across the sky (hemisphere), {where each sun position is perturbed by a small amount of random noise.} The goal is to uniformly represent direct solar illumination from all possible directions, not necessarily representing a set of realistically observable illumination directions. See Figure~\ref{fig:dataset_illumination}(a) for a demonstration of this definition of illumination directions. Given this set of Sun positions for a particular DEM patch, a set of images is rendered using point source solar illumination. During training with this dataset, the entire set of 30 images is provided for each DEM patch for SR. The purpose of training using this dataset is to initially orient the model, giving a sense of all possible illumination directions and how to relate topography to optical information. Additional fine-tuning of models to more realistic scenarios is conducted using the dataset described in the next subsection.

\subsection{Realistic Illumination Dataset}\label{sec:realistic_illum}
The set of illumination directions for a particular DEM patch in the Fibonacci lattice dataset is not generally realistically observable from a fixed location on the lunar surface. For a fixed location (longitude-latitude), the Sun can only be positioned within a solid-angle ``band'' across the sky, above the horizon (which we take into account when raytracing, along with neighboring topography surrounding each DEM patch). We make a simple assumption: given the position of the patch, the illumination direction is bound to an arc. The DEM patch can be associated with a set of realistically-illuminated images whose illumination directions are "sufficiently close" to this arc. In practice, during training, we define the longitude/latitude of each patch randomly during each pass over the dataset, such that each patch will have a different set of realistic (feasible) images each time it is considered during training. Therefore, for each patch, we require a large set of rendered images corresponding to a sufficiently dense set of illumination directions.

For all DEM patches, the illumination directions of the rendered images are at fixed elevations of 5{\textdegree}, 10{\textdegree}, 20{\textdegree}, 30{\textdegree}, 45{\textdegree}, 60{\textdegree}, 75{\textdegree}, and 85{\textdegree} from the horizon. For the lower elevations (5,10,20,30 {\textdegree}), illumination directions are defined every 5{\textdegree} in azimuth (72 ``longitudinal'' sun positions per elevation ``ring''). The remaining rings at elevations 45{\textdegree}, 60{\textdegree}, 75{\textdegree}, and 85{\textdegree} consist of illumination directions with azimuth spacings of 10{\textdegree}, 20{\textdegree}, 30{\textdegree}, and 60{\textdegree}, respectively. Each DEM patch is thus rendered a total of 360 times (once per illumination direction). At training and evaluation time, a subset of these rendered images is considered ``realistically observable'' depending upon the patch's randomly selected location on the lunar surface. Figure~\ref{fig:dataset_illumination}(b) demonstrates the definition of the valid illumination angle arc, which corresponds to a set of rendered images. This arc depends upon the longitude and latitude of the DEM patch.

\begin{figure}[!t]
    \centering
    \includegraphics[width=\linewidth]{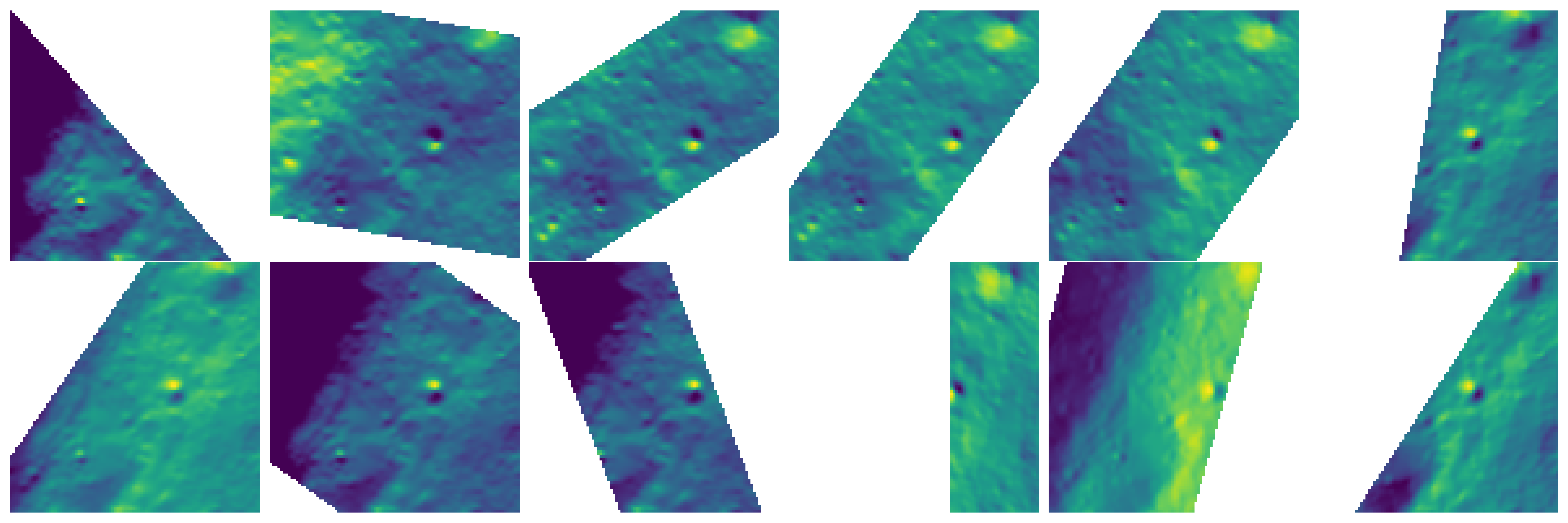}
    \caption{Missing data in rendered images emulates realistic observation patterns in NAC optical images. A band of observed pixels is defined in each fully rendered image, and pixels outside of this range are discarded. This set of images represents a partial observation of the same setting displayed in the middle panel of Figure~\ref{fig:dataset_illumination}.}
    \label{fig:missing_data}
\end{figure}

\subsection{Missing Optical Data}\label{sec:missing_data}
In many realistic settings, individual optical data products will not fully capture the region-of-interest (the DEM patch). Using a fixed viewing window defined by the DEM patch coordinates, this phenomenon can be defined as missing pixels in an individual image. The ability of SR methods to handle missing data is therefore crucial for application to realistic settings. To address such scenarios during training, before an image is provided to the model, a slice of ``observed'' pixels is determined. NAC-like measurement patterns can be represented by a  ``band'' of observations. To emulate such observations in rendered imagery, for a given image, a pair of parallel lines are selected (with some randomly-determined distance), and the pixels outside of this band are discarded. In Figure~\ref{fig:missing_data}, typical observed data patterns for rendered image patches are shown. During training and evaluation, missing pixels are encoded by setting their pixel values to -1. The intuition is that a model trained with such data will learn to ignore regions of images that are missing.

\section{{Training and Evaluation}}\label{sec:experiment}

{
In this section, we outline the parameterization and training of a SB model to solve the posterior sampling problem of Equation~\ref{eq:sr_problem}. That is, given low-resolution \textit{a priori} topography and high-resolution image context, a SB model is used to produce high-resolution topography reconstructions (see Section~\ref{sec:topo_sr_sb} for model details). This is accomplished via training and evaluation on the rendered topography datasets of Section~\ref{sec:datasets}.
}

To produce reconstructions for a given DEM, a single sample can be drawn from the pre-trained SB model. However, due to the stochasticity of the sampling scheme, multiple samples can be drawn to compute averages and standard deviations per pixel; the latter is denoted modeling uncertainty $\hat{\sigma}$. The RMS modeling uncertainty can then be compared to the true RMS error as an assessment of the accuracy of uncertainty quantification. Evaluation metrics can be computed on either a single sample from the SB or on the (pixel-wise) mean aggregation of samples. This is especially important and valuable for scientific applications, as well-calibrated, realistic error estimates for higher-resolution topographic models are critical to their use. 

Sets of reconstructed DEM patches are evaluated according to a range of metrics. Elevation error can be quantified by the root-mean-square (RMS) elevation error, as well as the 75th and 95th percentiles of the elevation error distribution. To evaluate first-order features, we compute the average slope errors in the vertical and horizontal directions. To evaluate higher-order, ``semantic'' features, the Fr{\'e}chet Inception Distance (FID) is used~\citep{heusel2017gans}, which is a common metric used for image evaluation in computer vision tasks. The Fr{\'e}chet distance reflects the difference between two probability distributions (as opposed to two individual samples). FID compares the distribution of Inception-v3 output maps of the true data to the distribution of Inception-v3 output maps of the reconstructions, {where Inception-v3 is a pre-trained convolutional neural network (CNN)~\citep{szegedy2015going} commonly used for computer vision evaluation metrics.} Intuitively, this {extracts and} evaluates features of the reconstructed data {using pre-trained neural networks}, assessing their fidelity with respect to the true distribution of high-resolution DEM patches.

\subsection{Model Parameterization}\label{sec:parameterization}
In this sub-section, we outline the parameterization details for each component of the SB framework. First, we define a low-dimensional latent representation space for topography data. An SB is defined in this latent space, and the dynamics of this SB are modeled by a neural network.

\textbf{Latent Space Parameterization.}
We train a VAE to embed each high-resolution DEM patch $x\in\mathbb{R}^{96\times96}$ into a low-dimensional representation $z\in\mathbb{R}^{4\times12\times12}$. Following \cite{rombach2022high}, the VAE is trained using a mixed objective to minimize reconstruction error and an adversarial objective. To parameterize the VAE networks, we use a residual CNN encoder $\mathcal{E}^\phi$ and decoder $\mathcal{D}^\phi$ (parameterized by $\phi$), similar to the encoder/decoder arms of a U-Net. See Appendix~\ref{app:vae} for a detailed description of the VAE neural network architecture and optimization. 

\textbf{Schr{\"o}dinger Bridge Parameterization.}
We define an SB in the latent space of the topography VAE. In this case, $p_\calB$ represents the distribution of latent-space \textit{a priori} DEMs, which is mapped to that of target-resolution DEMs, $p_\calA$. Following \cite{liu2023} for the parameterization of SB SDEs, we set $f_t=0$, {and $\beta_t$ symmetrically peaks at a maximum of $\beta_{T/2}=\beta_{\rm max}$ from $\beta_0=\beta_T=\beta_{\rm min}$:}
\begin{equation*}
    {
    \beta_t = \frac{2(\beta_{\rm min} - \beta_{\rm max})}{T} \left| t - \frac{T}{2} \right| + \beta_{\rm max},
    }
\end{equation*}
{where $\beta_{\rm max}=3\times10^{-4}$, $\beta_{\rm min}=10^{-4}$, and $T=1000$.} The SB connects the distribution of high-resolution latent-space DEMs, $z_0:=\mathcal{E}^\phi(x_0)\sim p_\calA$, to that of low-resolution latent-space DEMs, $z_T:=\mathcal{E}^\phi(x_T)\sim p_\calB$. As in Equation~\ref{eq:sb_sde_forward_linear}, the forward-time SB SDE dynamic has the following form:
\begin{equation}\label{eq:latent_forward_sde}
    \diff z_t = \sqrt{\beta_t}\diff w_t,\quad z_0\sim\hat{\Psi}_0,
\end{equation}
where $\hat{\Psi}_0=\delta_a$ for some $a\sim p_\calA$.

\textbf{Sampling Dynamic Parameterization.}
Given the VAE and a pre-defined SB, we can parameterize and learn the dynamics of the reverse process (low-resolution to high-resolution):
\begin{equation}\label{eq:latent_reverse_sde}
    \diff z_t = - \beta_t \nabla_{z_t} \log \hat{\Psi}_t(z_t) \diff t + \sqrt{\beta_t}\diff\bar{w}_t,\quad z_T\sim p_\calB.
\end{equation}
At each step along the sampling trajectory, a neural network $u_t^\theta$ receives $z_t$ (the partially-recovered latent representation) as input and yields an approximation of $\nabla_{z_t}\log\hat{\Psi}_t(z_t)$, which can be used to simulate a step of the SDE. The sampling process is initialized at time $T$ with the VAE encoding of a low-resolution DEM patch $x_T$ bicubic-upsampled to the target resolution. In addition to taking the time $t$ as conditional input, the network receives the low-resolution DEM $x_T$ and the set of rendered images $\mathcal{Y}$ as conditional input at each step of the sampling process to constrain the process towards generating \textit{true} high-resolution topography. The dynamics are approximated using a ViT with convolutional encoders for conditional inputs; see Figure~\ref{fig:sb_flowchart} for a visualization of the SB parameterized by a ViT and see Appendix~\ref{app:sb_vit} for a detailed description of the SB neural network architecture. In practice (and in all of the experiments outlined in this work), the sampling process can be conducted in as few as 10 steps simulating the SDE Equation~\ref{eq:latent_reverse_sde}.

\begin{figure}[t!]
    \centering
    \includegraphics[width=0.95\textwidth]{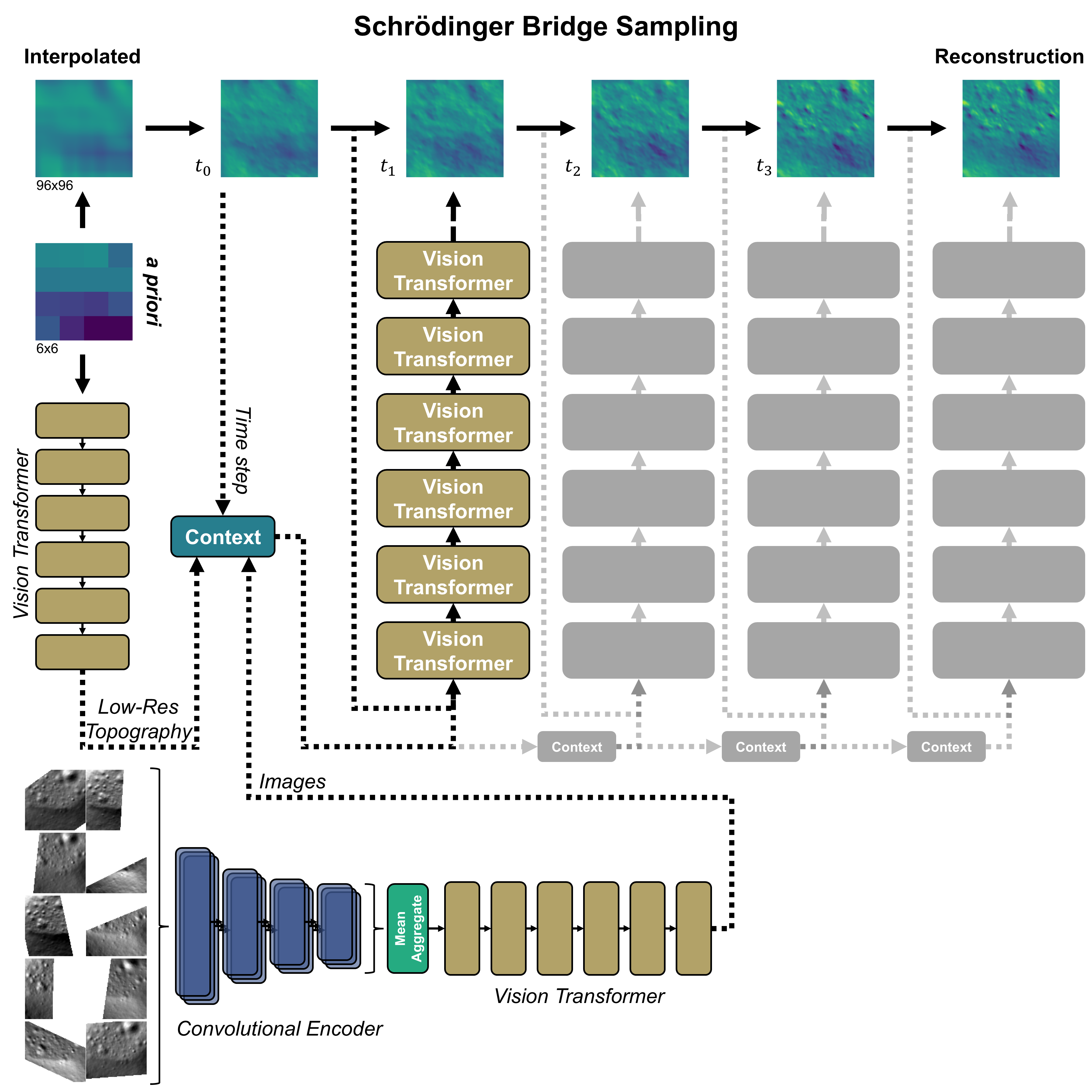}
    \caption{Flow chart of the SB generative process. The main ViT takes the partially-generated patch from the previous time step as input to yield the next-time-step patch (see Appendix~\ref{app:vae} for details of the VAE). Three pieces of conditional information are incorporated: the encoded generative process time step, an embedding of the \textit{a priori} topography via another ViT, and an embedding of the set of images via a set convolutional encoder and a ViT. This conditional information (denoted ``context'') is processed by the ViT in each time step (see Appendix~\ref{app:sb_vit} for a detailed description of the network architecture).}
    \label{fig:sb_flowchart}
\end{figure}

\begin{figure}[t]
    \centering
    \includegraphics[width=\linewidth]{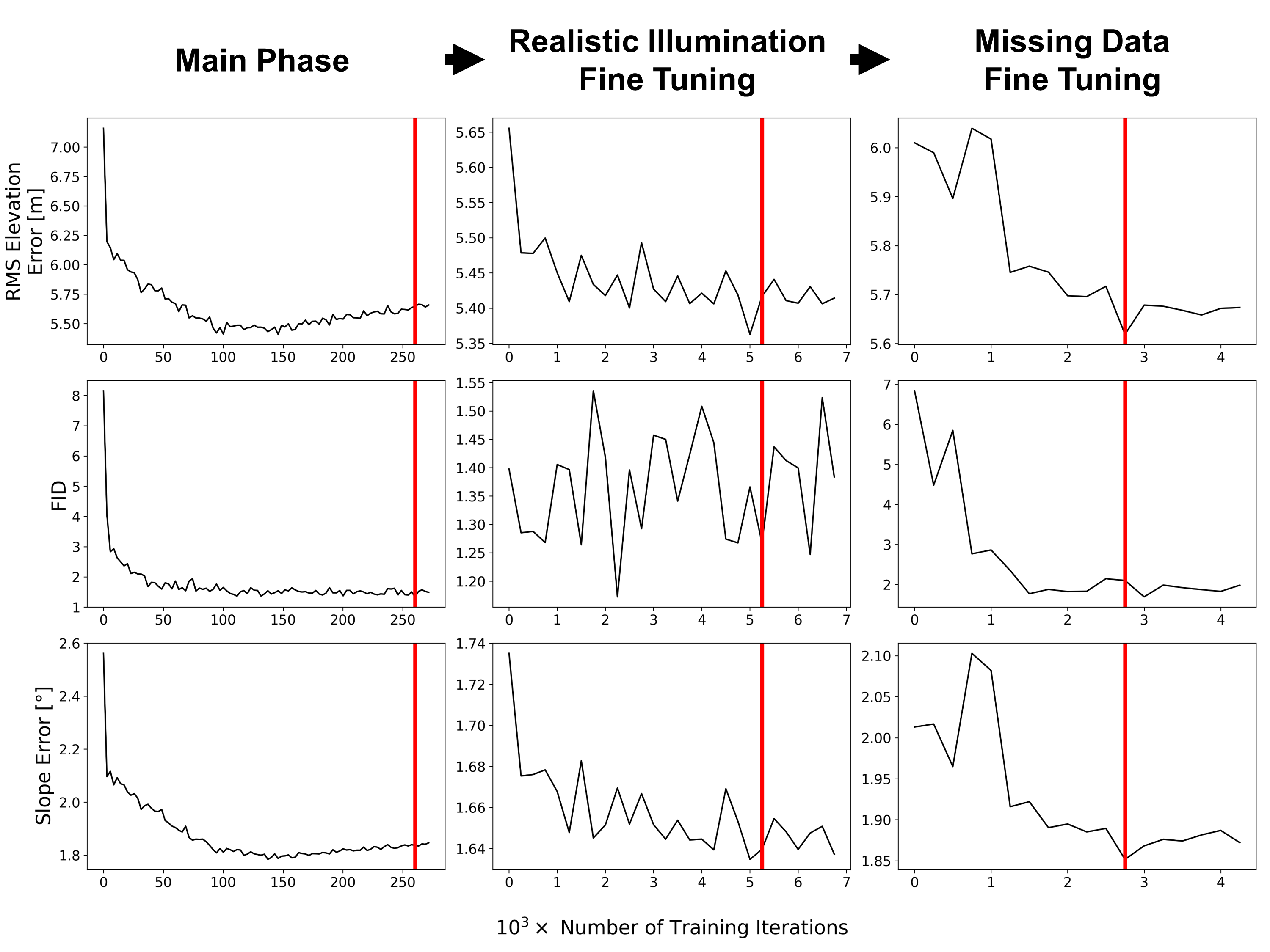}
    \caption{Evaluation metrics computed on 1,024 validation dataset patches throughout each phase of SB model training. The network parameters passed to the next phase are determined by selecting training iterations in which the validation metrics are small (vertical red lines). {Training in the first stage is halted when FID is smallest, which also corresponds to low elevation MSE and slope error.} Note the change in scale of the horizontal and vertical axes in the fine-tuning phases.}
    \label{fig:bridge_losses}
\end{figure}

\begin{table}[t]
    \caption{SB application to the test dataset provided 75 images with missing data per patch.}
    \label{tab:main_table}
    \centering
    \resizebox{\linewidth}{!}{
    \begin{tabular}{l|ccccccc}
        \hline
        \multirowcell{2}{Method} & \multirowcell{2}{Elev. \\ RMSE (m)} & \multirowcell{2}{Elev. Error \\ 75th Perc. (m)} & \multirowcell{2}{Elev. Error \\ 95th Perc. (m)} & \multirowcell{2}{RMS \\ $\hat{\sigma}$ (m)} & \multirowcell{2}{FID} &  \multirowcell{2}{Horiz. Slope \\ Error (\textdegree)} &   \multirowcell{2}{Vert. Slope \\ Error (\textdegree)} \\ \\
        \hline
        Bicubic Interpolation & 6.79 & 5.36 & 13.7 & - & 16.3 & 2.32 & 2.33 \\
        \hline
        Main Phase (1 sample) & 8.87 & 8.60 & 17.8 & - & 12.0 & 2.44 & 2.37 \\
        Main Phase (20 samples) & 8.22 & 7.68 & 16.6 & 3.39 & 12.6 & 2.26 & 2.19 \\
        \hline
        Illum. Fine Tune (1 sample) & 6.97 & 7.39 & 13.8 & - & 12.4 & 2.76 & 2.75 \\
        Illum. Fine Tune (20 samples) & 4.63 & 4.47 & 9.18 & 5.36 & 6.63 & 2.00 & 1.97 \\
        \hline
        Missing Fine Tune (1 sample) & 5.64 & 6.00 & 11.2 & - & 0.93 & 1.82 & 1.87 \\
        Missing Fine Tune (20 samples) & 3.56 & 3.53 & 6.99 & 4.50 & 5.42 & 1.46 & 1.50 \\
        \hline
        \multicolumn{8}{l}{$^{a}$Abbreviations denote elevation (Elev.), percentiles (Perc.), horizontal (Horiz.), and vertical (Vert.).}
    \end{tabular}
    }
\end{table}

\begin{figure}[t]
    \centering
    \includegraphics[width=\linewidth]{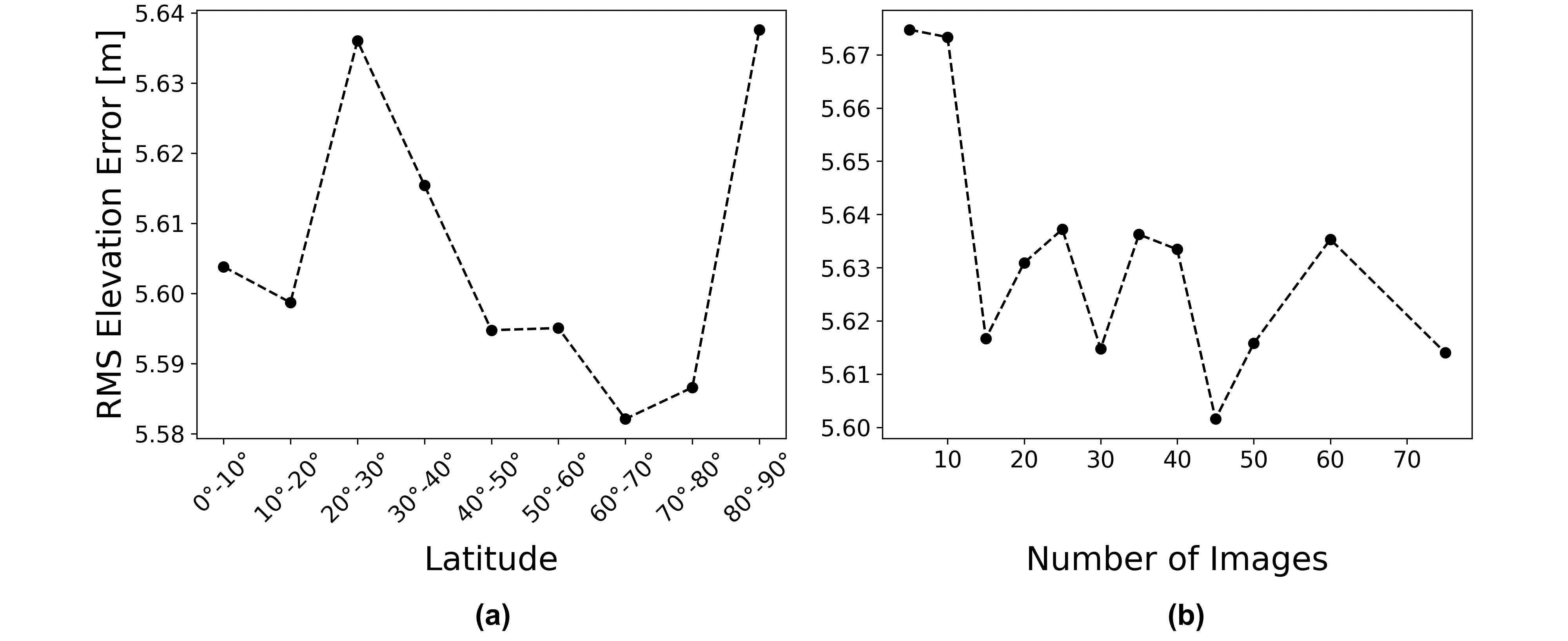}
    \caption{Elevation error of single-sample reconstructions on the 9,171 test dataset patches from the final fine-tuned model as a function of planetary latitude (a) and number of images (b). {There is no significant relationship between elevation error with respect to latitude or the number of images.}}
    \label{fig:perf_eval}
\end{figure}

\begin{figure}[t!]
    \centering
    \includegraphics[width=0.7\linewidth]{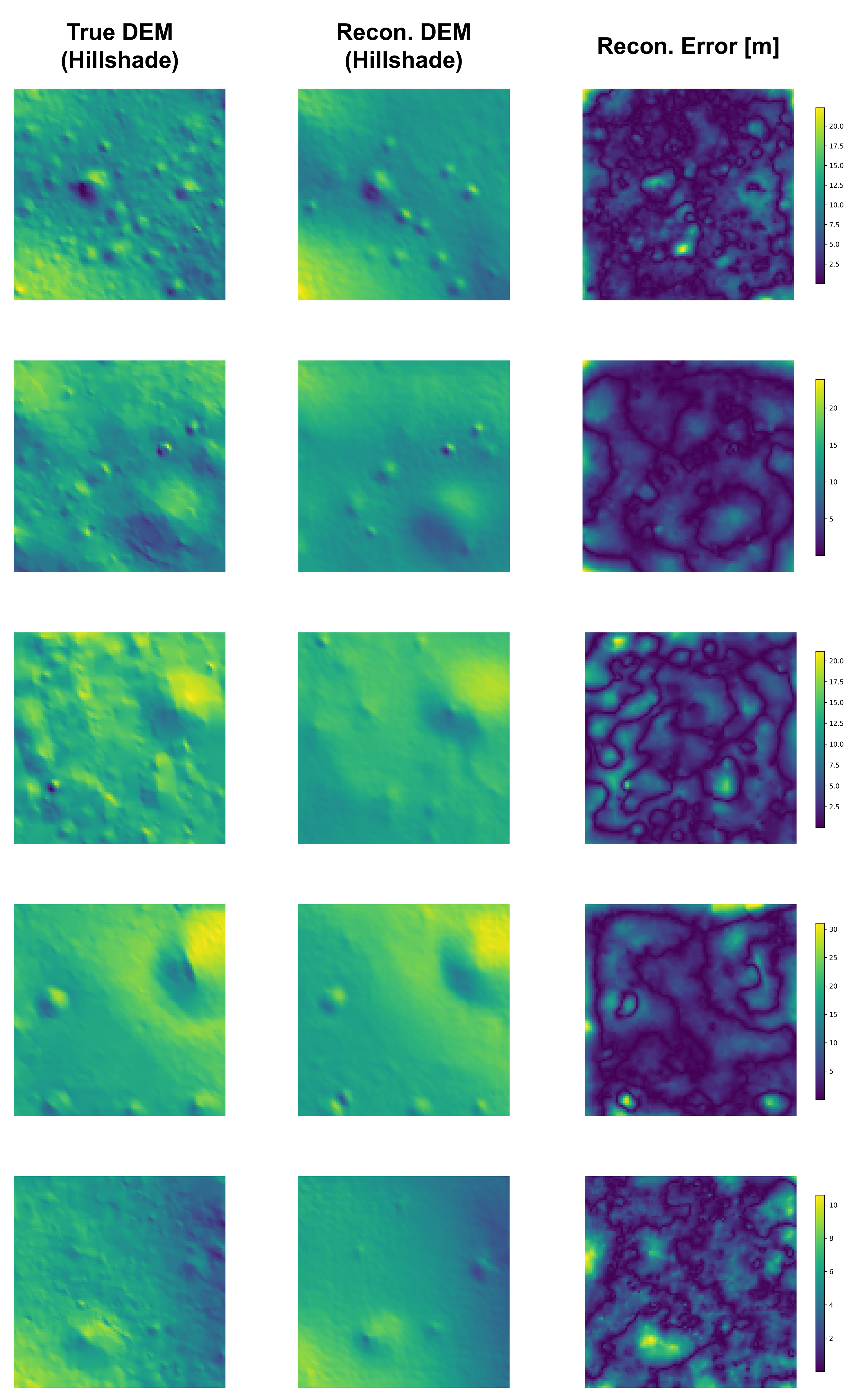}
    \caption{The final fine-tuned SB model is applied to 5 DEMs patches (96 $\times$ 96 pixels at 20 mpp) when providing 5 images per DEM patch. Reconstructions are aggregated over 20 samples.}
    \label{fig:recon_panels}
\end{figure}

\begin{figure}[t!]
    \centering
    \includegraphics[width=\linewidth]{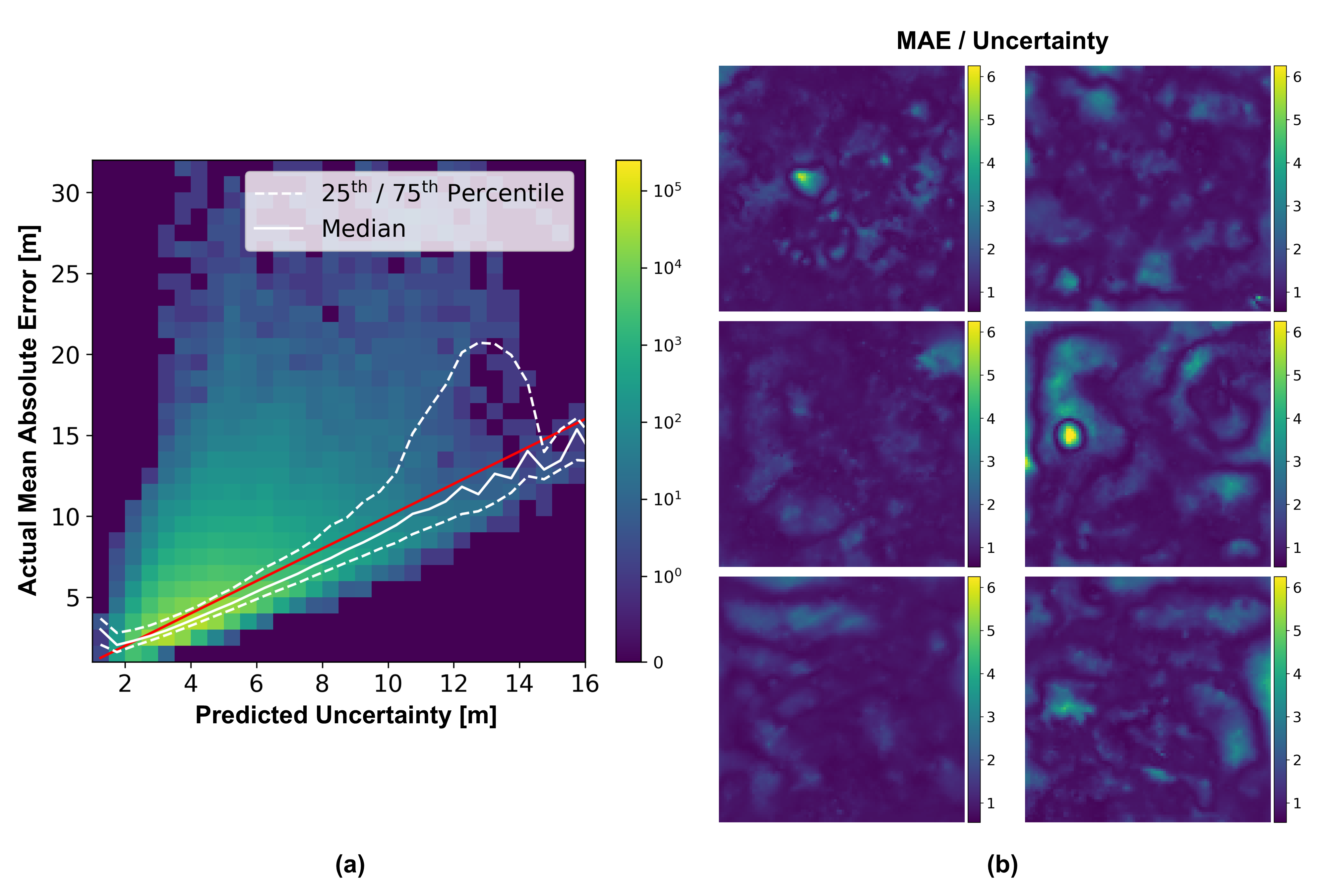}
    \caption{In (a), the correlation between prediction uncertainty and MAE is visualized, with 1:1 line shown in red and quantiles of the column marginals shown as white lines. In (b), the ratio between the MAE and the prediction uncertainty is shown for six different test dataset patches. {We find that the predicted uncertainty is typically within a factor of $\sim$2 of the true elevation MAE.}}
    \label{fig:err_v_unc}
\end{figure}

\subsection{Model Training}
In this section, we describe how neural network parameters are learned using the training datasets. The SB neural network parameters are learned via minimization of the DSM objective (Equation~\ref{eq:dsm}) corresponding to Equation~\ref{eq:latent_forward_sde}. Given high-resolution DEMs, their low-resolution counterparts, and corresponding (rendered) imagery, an empirical form of the DSM objective can be computed at randomly-sampled times $t$. The objective is minimized with respect to the ViT parameters $\theta$ using AdamW~\citep{loshchilovdecoupled}, a stochastic-gradient-descent-like first-order numerical optimization technique. The details of model training are outlined below.

The models are trained using the set of lunar DEM patches described in Section~\ref{sec:datasets}, with additional augmentations, including random flips along the vertical and/or horizontal axes. These data are split into 80\% (training), 10\% (validation), and 10\% (testing) partitions. First, the VAE parameters are learned using the VAE adversarial objective (Equation~\ref{eq:vae_loss}) computed on the training partition. Using the Adam optimizer \citep{kingma2014adam} with a batch size of 64, a learning rate of $4.5\times10^{-6}$, betas 0.5 and 0.9, and otherwise default PyTorch parameters, the VAE was trained for approximately 1.94 million iterations, or 1,690 epochs (passes over the training dataset). See Appendix~\ref{app:vae} for further details regarding the VAE optimization.

The latent-space SB is trained using rendered data corresponding to the same 80\% training partition used to train the VAE. The main phase of SB training is conducted using the Fibonacci lattice rendered data described in Section~\ref{sec:fibonacci}. The intuition behind this choice is that the majority of the optimization time should focus on encoding lighting information for the entire spectrum of illumination conditions. Afterwards, two additional phases of ``fine-tuning'' optimization are conducted, initializing the weights of the neural network with those at the end of training from prior phases. First, the model is fine-tuned using the realistic illumination conditions described in Section~\ref{sec:realistic_illum} with no missing data. Additionally, each DEM patch is associated with a random number of images (from 5 to 30){, such that each pixel will typically be non-shadowed in at least a few complementary images.} Second, the model is further fine-tuned using the same realistic illumination conditions and a potentially larger random number of images (5-100 images) with the missing data pattern described in Section~\ref{sec:missing_data}, where each image has a missing-data region.

Given a set of $N$ training points $\left( x^{(i)}_0,x^{(i)}_T, \mathcal{Y}^{(i)} \right)_{i=1}^N $, the SB neural network can be optimized numerically. Specifically, the DSM minimization (Equation~\ref{eq:dsm}) can be conducted using an empirical version of the loss~\citep{liu2023}:
\begin{equation}\label{eq:empirical_dsm}
     \theta^\star = \underset{\theta}{\text{arg min}} \ \sum_{i=1}^N \ \left\| u_{t^{(i)}}^\theta\left(z^{(i)}_{t^{(i)}},x_T^{(i)},\mathcal{Y}^{(i)}\right) - \frac{z^{(i)}_{t^{(i)}}-z^{(i)}_0}{\sigma_{t^{(i)}}} \right\|,
\end{equation}
such that $u_t^{\theta^\star}(z_t,\cdot,\cdot)\approx\nabla_{z_t} \log \hat{\Psi}_t(z_t)$. In Equation~\ref{eq:empirical_dsm}, each ${t^{(i)}}\sim\text{Uniform}(0,1)$ identically and independently, and $z^{(i)}_{t^{(i)}}\sim q\left(z^{(i)}_{t^{(i)}} \left| z^{(i)}_0, z^{(i)}_T\right.\right)$, an analytic Gaussian posterior (Equation 11 of \cite{liu2023}). This minimization is conducted over mini-batches of the training dataset using AdamW with a batch size of 256, a learning rate of $10^{-4}$, zero weight decay, and otherwise default PyTorch parameters.

SB model selection was conducted by identifying the stage of training with the best (lowest) evaluation metrics of the model reconstruction on a subset ($1,024$ patches) of the validation dataset. The model selected after the first phase of training (on Fibonacci lattice illumination angles) was trained for $260,260$ iterations. The first phase of fine-tuning (realistic illumination) was conducted for $5,250$ iterations, and the second phase (missing data) was conducted for $2,750$ iterations. Evaluation metrics applied to this subset of the validation dataset throughout training are visualized in Figure~\ref{fig:bridge_losses}. Notably, the fine-tuning phases do not require as many iterations of optimization to converge. This validates the intuition that the majority of information was learned by the model in the main phase of training.

\subsection{Model Evaluation}
The trained and fine-tuned SB models are evaluated at 16$\times$ SR (320 mpp to 20 mpp) on the testing partition of the LOLA DEM patches dataset (Section~\ref{sec:datasets}) which consists of $9,171$ patches. 75 images with NAC-like missing data patterns are provided for each patch, emulating realistically-observable sets of NAC images. For each SB model, either a single sample is produced or 20 samples are averaged (at the pixel level) to produce an aggregated reconstruction. In Table~\ref{tab:main_table}, the validation metrics computed on the patches reconstructed using each approach are displayed. Since the main phase model and first fine-tuned model are not trained for this realistic setting, they perform similarly to the baseline interpolation in the elevation and slope metrics, but the FID is significantly lower. This can be interpreted as follows: these models produce more realistic-looking DEMs than interpolation, but they are not consistent with the true topography.

The fully fine-tuned model significantly outperforms interpolation and all prior models. Given a single sample from the SB, RMS elevation reconstruction error is 5-6 m, or 25\%-30\% of the pixel width. Additionally, slope errors are less than 2 degrees on average. Given 20 samples from the model, the elevation and slope errors decrease, with a slight increase in FID. This can be interpreted as a smoothing of the higher-order features due to aggregation of multiple samples, keeping the reconstruction consistent with elevation and first-order features. Note, however, that the FID of the aggregated reconstructions is still better than that of interpolation and previous models.

The generalization capacity of the final fine-tuned model is assessed in Figure~\ref{fig:perf_eval}, which displays the RMS elevation error of single-sample reconstructions on 1,000 patches from the test dataset under different conditions. In Figure~\ref{fig:perf_eval}(a), the 75 images provided for each patch correspond to patches at fixed ranges of latitudes; the results indicate that the performance does not degrade dramatically with latitude. In Figure~\ref{fig:perf_eval}(b), the number of images provided to the model is varied from 5 to 75, and the performance only degrades slightly given fewer images. A set of reconstructions given 5 images per patch (assuming each patch is located at 0\textdegree \ E and 60\textdegree \ N)  is shown in Figure~\ref{fig:recon_panels}. Consistent with the quantitative results, the aggregation of 20 samples smoothes some high-frequency features, potentially removing inaccurate small-scale topography and yielding an accurate overall reconstruction.

Figure~\ref{fig:err_v_unc} demonstrates the capacity of the uncertainty provided by our approach to capture the true reconstruction error. These results correspond to reconstructions of 50 test dataset patches (20 clones per patch) each provided with 30 images with missing data (assuming 0\textdegree \ E and 60\textdegree \ N). At each pixel, we compare the prediction uncertainty, i.e., the standard deviation of 20 predictions per pixel, to the mean absolute error (MAE), i.e., the average of 20 absolute reconstruction errors per pixel. In Figure~\ref{fig:err_v_unc}(a), the uncertainty can be seen to have an approximately linear relationship with the MAE; Pearson's $r$ is 0.63, indicating a weak linear correlation. This can be attributed to the off-linear heavy tail of the marginal distributions visualized in Figure~\ref{fig:err_v_unc}(a). {Despite this, the median MAE (solid white line) follows the 1:1 line (red) closely, suggesting that the predicted uncertainty is generally reasonably accurate.} Figure~\ref{fig:err_v_unc}(b) visualizes the ratio of MAE and prediction uncertainty, demonstrating that typical errors are on the order of at most a few $\hat{\sigma}$. In some cases, uncertainty underestimation can be attributed to pixels not captured by any of the images. However, we find that the uncertainty underestimates the MAE in some other, less explainable cases as well. In general, the MAE is typically largest at locations in which the uncertainty is also largest, which are intuitively the locations where reconstructions should be least trusted.

\begin{figure}[t!]
    \centering
    \includegraphics[width=\linewidth]{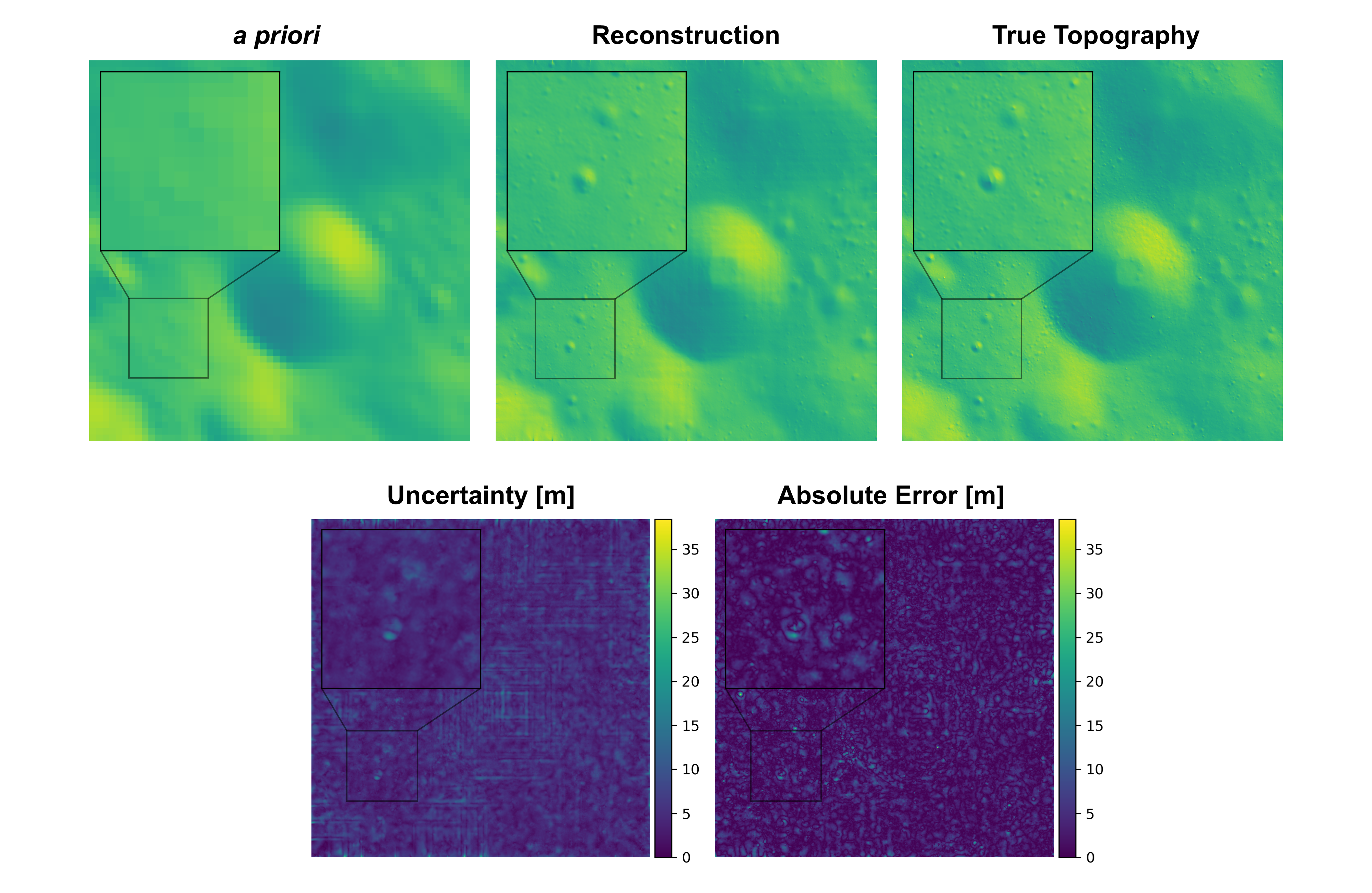}
    \caption{A large area (19.2 $\times$ 19.2 km) is reconstructed using 37 rendered images (corresponding to a feasible set of illumination directions for a patch at 0\textdegree \ E, 40\textdegree \ N). The reconstructed topography recovers true features not present in the \textit{a priori} topography and the uncertainty map roughly tracks the true absolute error.}
    \label{fig:large_rendered}
\end{figure}

\begin{figure}[t!]
    \centering
    \includegraphics[width=\linewidth]{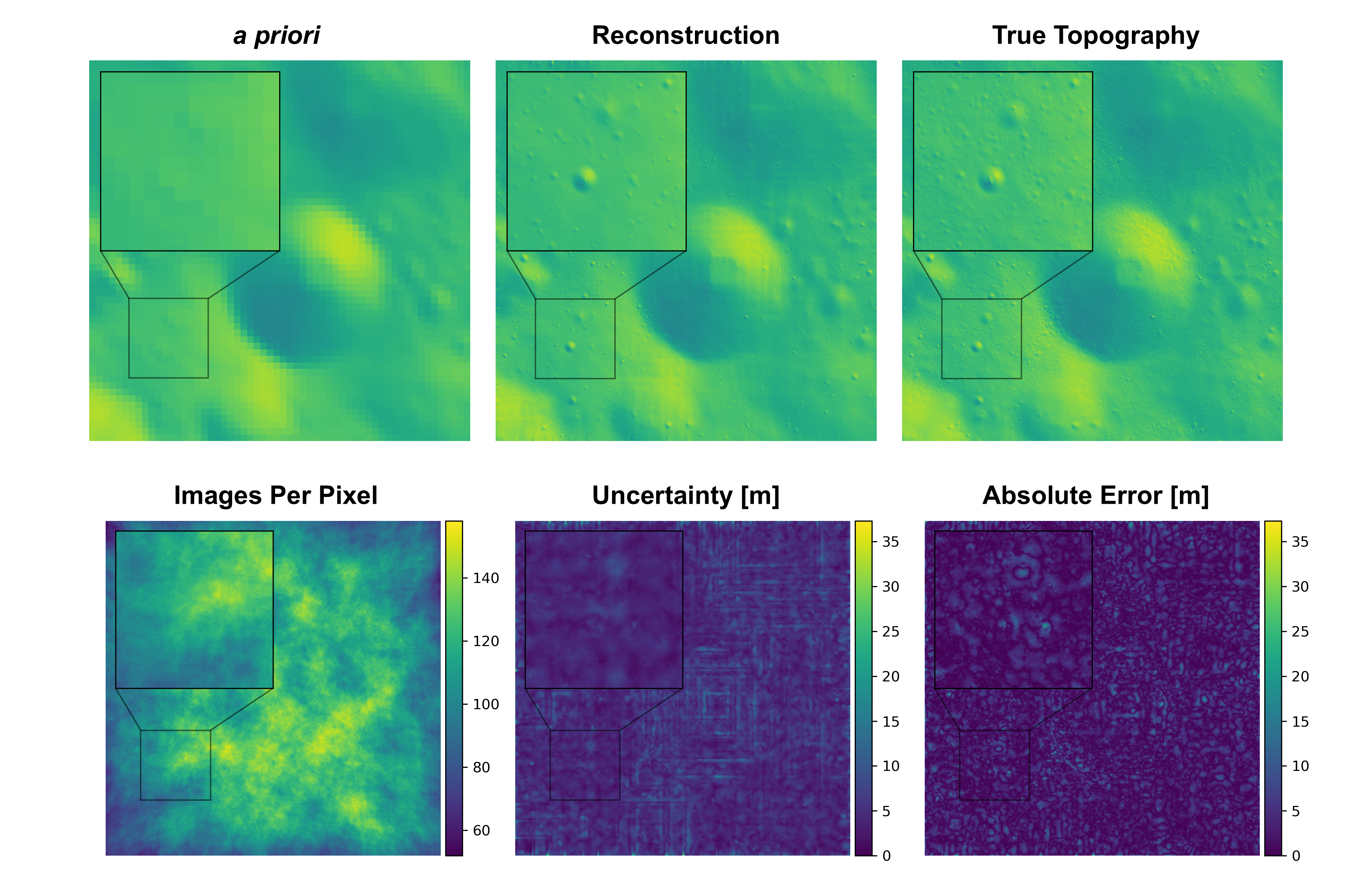}
    \caption{The same large area as in Figure~\ref{fig:large_rendered} (again assuming it is located at 0\textdegree \ E, 40\textdegree \ N) is reconstructed given 1,500 rendered images with NAC-like missing data patterns.}
    \label{fig:large_rendered_missing}
\end{figure}

\begin{figure}[t!]
    \centering
    \includegraphics[width=\linewidth]{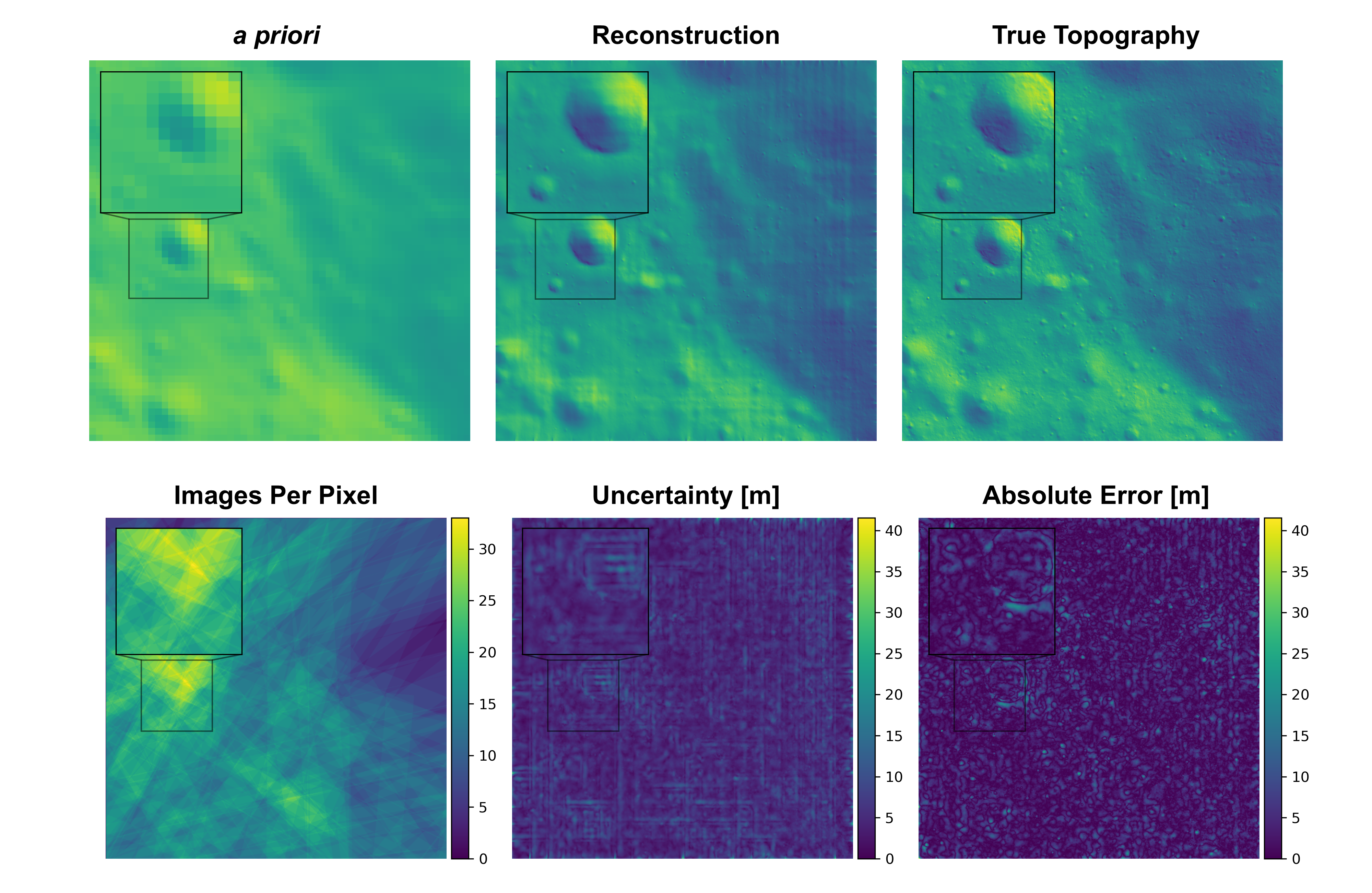}
    \caption{A region within the Faustini Rim A candidate landing region for the upcoming Artemis missions is reconstructed given 145 real NAC images.}
    \label{fig:large_real_nac}
\end{figure}

\section{Application to Large-Scale Topography}
\label{sec:application}
We outline a mosaicing scheme for the application of the SB method to areas larger than 96$\times$96 pixels. Our approach is similar to the {\texttt{sfs} and} \texttt{dem\_mosaic} functions of the Ames Stereo Pipeline (ASP)~\citep{beyer2018ames}, which {improve and} blend overlapping DEMs, respectively. {However, {unlike the \texttt{sfs} function of ASP, SB sampling allows for the computation of multiple predictions (clones) per patch, facilitating uncertainty analysis.}} The SB is applied as a moving window over a large region with some specified stride $s$, resulting in overlapping reconstructed patches. The reconstruction for each pixel over the entire large area is produced via a weighted average of the pixels in each patch, where each pixel's weight for a given reconstructed patch corresponds to its distance from the nearest edge of the patch. Specifically, a grassfire filter (weight is proportional to the square of the distance to the nearest edge) convolved with a Gaussian of width $\sigma_f$ is applied to each patch before aggregation. The weighted standard deviation of each pixel can be computed using these weights as well. Multiple samples (`clones', similarly to \textit{e.g.},~\cite{lemoine2014grail}) of each patch can be generated by the model to yield more robust predictions.

To represent a realistic scenario over a large area, we consider a region which is 19.2 $\times$ 19.2 km (960 $\times$ 960 pixels). This corresponds to a LOLA DEM patch centered at 0\textdegree \ E and 85\textdegree \ S. The 16$\times$ mean-downsampled DEM is provided as \textit{a priori}. We render ``rings'' of images (360 images total) as described in Section~\ref{sec:realistic_illum}. First, we apply the SB approach to this large area with realistic sets of rendered images without any missing data. That is, a realistic Sun trajectory is defined according to the assumed position of the region on the lunar surface. The SB method is applied with filter blur $\sigma_f=2.0$ and stride $s=960$ meters, i.e., 48 pixels so that each patch overlaps 50\% with adjacent patches. Four clones are produced for each patch. This reconstruction approach is applied at multiple assumed latitudes (with longitude 0\textdegree \ E), but we do not observe significant degradation in reconstruction based on the assumed location of the region. The reconstruction for the region when assumed to be located at 0\textdegree \ E, 40\textdegree \ N is shown in Figure~\ref{fig:large_rendered}.

To consider a more realistic scenario, we also consider missing data in the provided images. Additionally, assuming the region is located at 0\textdegree \ E, 40\textdegree \ N, $1,500$ images are provided to the model, which is approximately the number of real NAC images that could be available over a region at that latitude. Since each image includes missing data, multiple images illuminated from the same sun angle are provided. See Figure~\ref{fig:large_rendered_missing} for a visualization of the reconstruction, again using $\sigma_f=2.0$, stride $s=960$, 4 clones per patch, and a 320 mpp (60 $\times$ 60 pixels) \textit{a priori} LOLA DEM patch. In this case, the reconstruction is not as high-fidelity as that produced using a smaller set of ``full'' images (Figure~\ref{fig:large_rendered}), but small-length-scale features are still recovered. 

Finally, we consider large-area reconstruction given real NAC images. Specifically, we consider the topography of a 19.2 $\times$ 19.2 km region (960 $\times$ 960 pixels at 20 mpp) within the Faustini Rim A Artemis candidate landing region. We use 145 real NAC images (downsampled to 20 mpp) to inform the SB reconstruction from a 320 mpp downsampled LOLA DEM (60 $\times$ 60 pixels). The same filter as in the other examples is used in this case ($\sigma_f=2.0$ and stride $s=960$ meters) and 4 clones are again produced for each patch. The reconstruction is visualized in Figure~\ref{fig:large_real_nac}. The resulting DEM is not as high-quality as in the rendered data and there are more artifacts related to mosaicing, but the method still yields a reasonable high-resolution DEM, resolving smaller-scale features than the \textit{a priori}. Some artifacts can be due to the relatively basic mosaic procedure{, in addition to inherent differences between real and rendered imagery}.

\section{Conclusions}
\label{sec:conclusions}
We have developed a SB-based method for topography SR on the Moon, yielding high-resolution terrain reconstructions via a stochastic sampling scheme. The SB is parameterized by a ViT neural network, which also takes into account high-resolution optical imagery, physically constraining the generative process. This method yields multiple reconstructions for each input, permitting computation of statistics for each pixel, yielding mean reconstructions and uncertainties. On a holdout test dataset of rendered lunar topography patches at 20 mpp, the method achieves a reconstruction error of 5-6 m. We also demonstrate the scalability of this method to arbitrarily large areas via a tiling and mosaicing scheme.

Practical application of this method necessitates efficient and accurate application to large areas at high resolution. Our experiments reveal good performance in such cases using rendered optical imagery, with a final example application to downsampled real NAC imagery. Future work will focus on further fine-tuning to real NAC images and a focus on higher resolutions than 20 mpp. Specifically, since NAC images are 0.5-3 mpp in resolution, the next steps include training on renderings of real data at that resolution. This will require an extensive dataset of high-resolution topography data at that resolution, which does not currently exist. However, data can be gathered and generated using existing methods such as stereophotogrammetry and SfS.

{We emphasize that the developments highlighted in this work should not yet be directly compared with traditional SfS approaches - though potential advantages include speed (after training) and the ability to efficiently produce informative (accurate) uncertainties. A key advantage of traditional approaches is that they do not require training of a machine learning model, and can therefore operate at arbitrarily high resolutions. To effectively compare with such approaches at meter-level resolutions, a large training dataset of high-resolution topography with no artifacts would be required, which does not yet exist. At this stage, we demonstrate the potential for state-of-the-art deep learning tools in the area of lunar topography, and leave higher-resolution enhancements to future work.}

This work demonstrates the potential of generative modeling for improving the quality of remote sensing data, incorporating multiple such data sources. Current limitations include under-estimation of error in the uncertainty metric, which could be addressed by a more principled error estimation scheme, e.g., using normalizing flows. Additionally, more advanced conditioning mechanisms can be investigated to better incorporate optical imagery, potentially leading to improved modeling accuracy and uncertainty. Extensions will focus on the adaptation of this method to more realistic settings and datasets.

\section*{Acknowledgments}
MR was supported through the NASA internship program.
Support for this research was provided by NASA's Planetary Science Division Research Program, through ISFM work package Planetary Geodesy at Goddard Space Flight Center.
SB acknowledges support by NASA under award number 80GSFC24M0006.
YX is partially supported by an NSF CAREER CCF-1650913, NSF DMS-2134037, CMMI-2015787, CMMI-2112533, DMS-1938106, DMS-1830210, ONR N000142412278, and the Coca-Cola Foundation.
Resources supporting this work were provided by the NASA Center for Climate Simulation (NCCS) at Goddard Space Flight Center, in particular their Prism GPU cluster. The Python code used to train these neural networks is published on GitHub and Zenodo (\url{https://github.com/mrepasky3/topo_sr_schrodinger/}) with code adapted from the I2SB repository (\url{https://github.com/NVlabs/I2SB}, \cite{liu2023}), the DiT repository (\url{https://github.com/facebookresearch/DiT}, \cite{peebles2023scalable}), the guided diffusion repository (\url{https://github.com/openai/guided-diffusion}, \cite{dhariwal2021diffusion}), the latent diffusion repository (\url{https://github.com/CompVis/latent-diffusion}, \cite{rombach2022high}), and the taming transformers repository (\url{https://github.com/CompVis/taming-transformers}, \cite{esser2021taming}).

\section*{Contributions}

All authors contributed equally. MR wrote the first draft. All the co-authors edited the manuscript.

\textit{Software}{CGAL version 5.6.0 \citep{cgal},  
          Shadowspy version 0.1.0 \citep{shadowspy}, 
          python-flux \citep{potter2023fast},
          xarray version 2024.01.1 \citep{hoyer2017xarray,hoyer_2024_xarray},
          Matplotlib version 3.8.2 \citep{Hunter:2007,matplotlib2023},
          PyTorch version 2.1.2 \citep{paszke2019pytorch},
          CUDA version 11.8 \citep{farber2011cuda},
          Torch image models version 0.9.16 \citep{rw2019timm}
          }

\appendix

\section{Deep Learning \& Function Approximation}\label{sec:depp_learning}

Many modern generative models use U-Nets~\citep{ronneberger2015u} to approximate $u_t^\theta$ (the neural network weights), e.g., for diffusion models~\citep{ho2020denoising}, normalizing flows~\citep{lipman2022flow}, and SBs~\citep{liu2023}. U-Nets are a type of convolutional neural network (CNN) with residual connections~\citep{he2016deep}, meaning intermediate data embeddings are transmitted across layers of the network. CNNs repeatedly apply convolutional filters, which are typically small kernels (e.g., $3\times3$ or $5\times5$) with learnable components. After each layer, the spatial dimensions of input tensor data are downsampled or upsampled; U-Nets are a particular type of CNN that first downsample input data to a small spatial resolution (the encoding arm), followed by successive layers to re-size data (the decoding arm) to the original shape supplied to the encoding arm. Recent improvements include adaptive normalization~\citep{dhariwal2021diffusion} and attention~\citep{vaswani2017attention,ho2020denoising}. Time-dependence is encoded using the sinusoidal embedding from transformers~\citep{vaswani2017attention}.

More recently, vision transformers (ViTs) have challenged U-Nets as powerful backbones for generative modeling~\citep{peebles2023scalable,esser2024scaling}. \cite{dosovitskiy2020image} introduced ViTs by applying the attention mechanism~\citep{vaswani2017attention} to patched tensor data. That is, an input tensor of shape $C\times H\times W$ is ``patchified'' along the spatial dimensions ($H$ and $W$) to yield $L:=HW/P^2$ number of sub-tensor patches of shape $C\times P\times P$. Using convolutional layers, patches are embedded into vectors of size $D$, and the vectorized patches are unrolled into a sequence of shape $L\times D$. The sequence is then input to $M$ subsequent transformer blocks, each block applying multi-head attention. See Figure~\ref{fig:background}(c) for a visualization of ViT input and architecture.

Diffusion Transformers (DiTs) adapt ViTs for diffusion modeling~\citep{peebles2023scalable}. DiTs are modified ViTs, which include conditioning, facilitating time-dependence, and consideration of contextual information. Since the score functions in diffusion and SBs depend on ``time'', effective generative modeling requires time-dependent neural networks. Moreover, dependence upon general conditional information $y$ can help constrain the generative process given context (that is, the score $u_t^\theta(x_t,y)$ can depend on $y$). To this end, DiTs incorporates an adaptive layer norm~\citep{perez2018film}, which derives ``neural network layer re-scaling parameters" from embeddings of time $t$ and conditional information $y$. In this work, we refer to ViTs and DiTs (which depend on conditional information) interchangeably.

Training generative models in high-dimensional settings (matrices and tensors) can be computationally expensive. Latent generative models~\citep{rombach2022high} alleviate this issue by learning the generative process in the \textit{latent space} of an autoencoder. Autoencoders are encoding/decoding neural networks which embed and decode high-dimensional data to and from a lower-dimensional latent space using learned compressive/decompressive mappings. For generative modeling, this neural network is trained before training the generative model neural network. In Variational Autoencoders (VAEs)~\citep{kingma2013auto}, the latent space is a low-dimensional Gaussian mixture distribution, and the projection into this space is learned by the encoder network. The decoder network reconstructs the original input data from its latent code. Provided a trained VAE, latent generative models learn a generative process in the latent space, such that generated latent codes can be decoded by the VAE into synthesized samples~\citep{rombach2022high}. See Figure~\ref{fig:background}(d) for a demonstration of generative modeling in the latent space of a VAE. VAEs used in generative models are typically residual CNNs that downsample tensor inputs.

\section{Variational Autoencoder Details}\label{app:vae}
In this section, we outline the optimization problem pertaining to the training of VAE neural network encoder/decoder $\mathcal{E}^\phi/\mathcal{D}^\phi$ parameterized by $\phi$. The first term of the VAE objective is a ``perceptual'' term $L_{\rm rec}$~\citep{zhang2018unreasonable} based upon a comparison between neural network embeddings of the input $x$ and reconstruction $\tilde{x} := \mathcal{D}(\mathcal{E}(x))$. An adversarial term $L_{\rm adv}$ is defined~\citep{esser2021taming} such that a discriminator $D^\psi$ (parameterized by $\psi$) is optimized alongside $\mathcal{E}^\phi/\mathcal{D}^\phi$ to assess the quality of the reconstruction $\tilde{x}$. Following \cite{kingma2013auto}, we regularize the distribution of the latent space representation $z:=\mathcal{E}^\phi(x)$ using a Kullback-Leibler divergence penalty $L_{\rm reg}$ between the latent space distribution and the standard normal distribution. The full objective has the following general form:
\begin{equation}\label{eq:vae_loss}
    L_{\rm VAE} = \underset{\phi}{\text{min}} \ \underset{\psi}{\text{max}} \left( L_{\rm rec}(x,\tilde{x}) - L_{\rm adv}(\tilde{x}) + \log D^\psi(x) + \rho_{\rm KL} L_{\rm reg}(z) \right),
\end{equation}
where $\rho_{\rm KL}=10^{-6}$ is the regularization weight following \cite{rombach2022high}.

Each network is composed of 4 successive layers, each of which consists of 2 residual CNN blocks followed by $\frac{1}{2}\times$ downsampling or $2\times$ upsampling of the spatial dimensions. The input is a ``tensor'' of 1 channel (i.e., a $1\times96\times96$ matrix), and the subsequent layers yield tensors of 128, 256, and 256 channels. That is, the shape of the embedded data after each layer of the encoder is $128\times48\times48$, $256\times24\times24$, and $256\times12\times12$. This is followed by a residual CNN block, self-attention, and another residual CNN block, maintaining the spatial size and number of channels. Finally, a convolution is applied to yield a representation including only 4 channels: $4\times12\times12$. The decoder is similarly structured, replacing downsampling with upsampling. The resultant combined encoder/decoder network consists of $\sim$21 million parameters. Following \citep{rombach2022high}, the adversarial network $D^\psi$ is defined by a convolutional neural network of 3 layers, consisting of approximately 2.8 million parameters.

In the modeling described in Section~\ref{sec:method}, the VAE and discriminator are optimized numerically using an empirical form of Equation~\ref{eq:vae_loss}. This is conducted in an alternating manner, where each network is optimized alternatively for 5 iterations each. However, for the parameters of the discriminator are fixed (at the random initialization) for the 5,000 iterations, to ``warm start'' the parameters of the VAE encoder/decoder.

\section{Schr{\"o}dinger Bridge Vision Transformer Architecture}\label{app:sb_vit}
In each time step of the SB generative process, the main body of the ViT takes a latent-space tensor of shape $4\times12\times12$ as input; the output is used to construct the latent space tensor in the next time step. This tensor is split into 36 patches of size $4\times2\times2$, each of which are embedded as vectors to yield a length-36 sequence of length-768 vectors. This sequence is passed through 6 transformer blocks with 12 attention heads per block; this is represented by the middle series of ViT blocks in Figure~\ref{fig:sb_flowchart}. In each transformer block, conditional information in the form of a length-768 vector (see the next paragraph) is infused using adaptive layer normalization~\citep{peebles2023scalable}.

Three sources of conditional information are encoded as context into a single length-768 vector embedding. The process of encoding contextual information is described below.

\

\noindent \textbf{Time step Context:} The time step $t$ of the generative process is encoded using transformer position embeddings~\citep{vaswani2017attention} into a length-768 vector.

\

\noindent \textbf{Image Set Context:} Each image is passed through a convolutional encoder with the same architecture as that of the VAE encoder (see Appendix~\ref{app:vae}), yielding a tensor embedding of shape $4\times12\times12$ for each image (bottom left of Figure~\ref{fig:sb_flowchart}). The set of image embeddings are averaged along each pixel to yield a single $4\times12\times12$ embedding for the entire set. Then, this tensor is patchified into 36 non-overlapping patches, each of size $4\times2\times2$, which are embedded as a length-36 sequence of length-768 vectors. This sequence is input to 6 subsequent transformer blocks with 12 attention heads per block to finally yield a single length-768 vector representation for the image set (bottom center of Figure~\ref{fig:sb_flowchart}).

\

\noindent \textbf{Low-Resolution Topography Context:} The low-resolution DEM patch $x_T$ is patchified into 9 patches of shape $2\times2$ pixels which are then encoded into a length-9 sequence of length-768 vectors. This sequence is passed through 6 transformer blocks of 12 attention heads each, again yielding a single length-768 vector (left side of Figure~\ref{fig:sb_flowchart}).

\

\noindent These three length-768 vector embeddings (time, images, and \textit{a priori}) are summed along each pixel to yield a single length-768 vector containing all conditional information (i.e., the ``context'').

\bibliography{arxiv_final}

\end{document}